\pdfoutput=1
\documentclass{article}

\usepackage{arxiv}

\usepackage[utf8]{inputenc} 
\usepackage[T1]{fontenc}    
\usepackage{hyperref}       
\usepackage{url}            
\usepackage{booktabs}       
\usepackage{amsfonts}       
\usepackage{nicefrac}       
\usepackage{microtype}      
\usepackage{lipsum}
\usepackage{graphicx}
\usepackage{algorithm}

\usepackage{algorithm}
\usepackage{algpseudocode}

\usepackage{graphicx}
\usepackage{wrapfig}

\usepackage{orcidlink}

\title{FruitPAL: An IoT-Enabled Framework for Automatic Monitoring of Fruit Consumption in Smart Healthcare}

\author{
\orcidlink{0009-0001-5516-4323}Abdulrahman Alkinani \\
Department of Computer Science and Engineering\\
University of North Texas \\
Denton, TX 76207 \\
\texttt{AbdulrahmanAlkinani@my.unt.edu}
\And
\orcidlink{0000-0002-8796-4819} Alakananda Mitra \\ 
Nebraska Water Center\\
University of Nebraska-Lincoln \\
Lincoln, NE 68588 \\
\texttt{amitra6@unl.edu}
\And
\orcidlink{0000-0003-2959-6541}Saraju P Mohanty \\
Department of Computer Science and Engineering \\
University of North Texas \\
Denton, TX 76207 \\
\texttt{saraju.mohanty@unt.edu}
\And
\orcidlink{0000-0002-1616-7628}Elias Kougianos \\
Department of Electrical Engineering \\
University of North Texas \\
Denton, TX 76207 \\
\texttt{elias.kougianos@unt.edu}
}

\begin{document}
\maketitle
\begin{abstract}
Fruits are rich sources of essential vitamins and nutrients that are vital for human health. This study introduces two fully automated devices, FruitPAL and its updated version, FruitPAL 2.0, which aim to promote safe fruit consumption while reducing health risks. Both devices leverage a high-quality dataset of fifteen fruit types and use advanced models—YOLOv8 and YOLOv5 V6.0—to enhance detection accuracy. The original FruitPAL device can identify various fruit types and notify caregivers if an allergic reaction is detected, thanks to YOLOv8’s improved accuracy and rapid response time. Notifications are transmitted via the cloud to mobile devices, ensuring real-time updates and immediate accessibility. FruitPAL 2.0 builds upon this by not only detecting fruit but also estimating its nutritional value, thereby encouraging healthy consumption. Trained on the YOLOv5 V6.0 model, FruitPAL 2.0 analyzes fruit intake to provide users with valuable dietary insights. This study aims to promote fruit consumption by helping individuals make informed choices, balancing health benefits with allergy awareness. By alerting users to potential allergens while encouraging the consumption of nutrient-rich fruits, these devices support both health maintenance and dietary awareness.
\end{abstract}


\section{INTRODUCTION}
\begin{figure}[htbp]
\centerline{\includegraphics[width=0.7\textwidth]{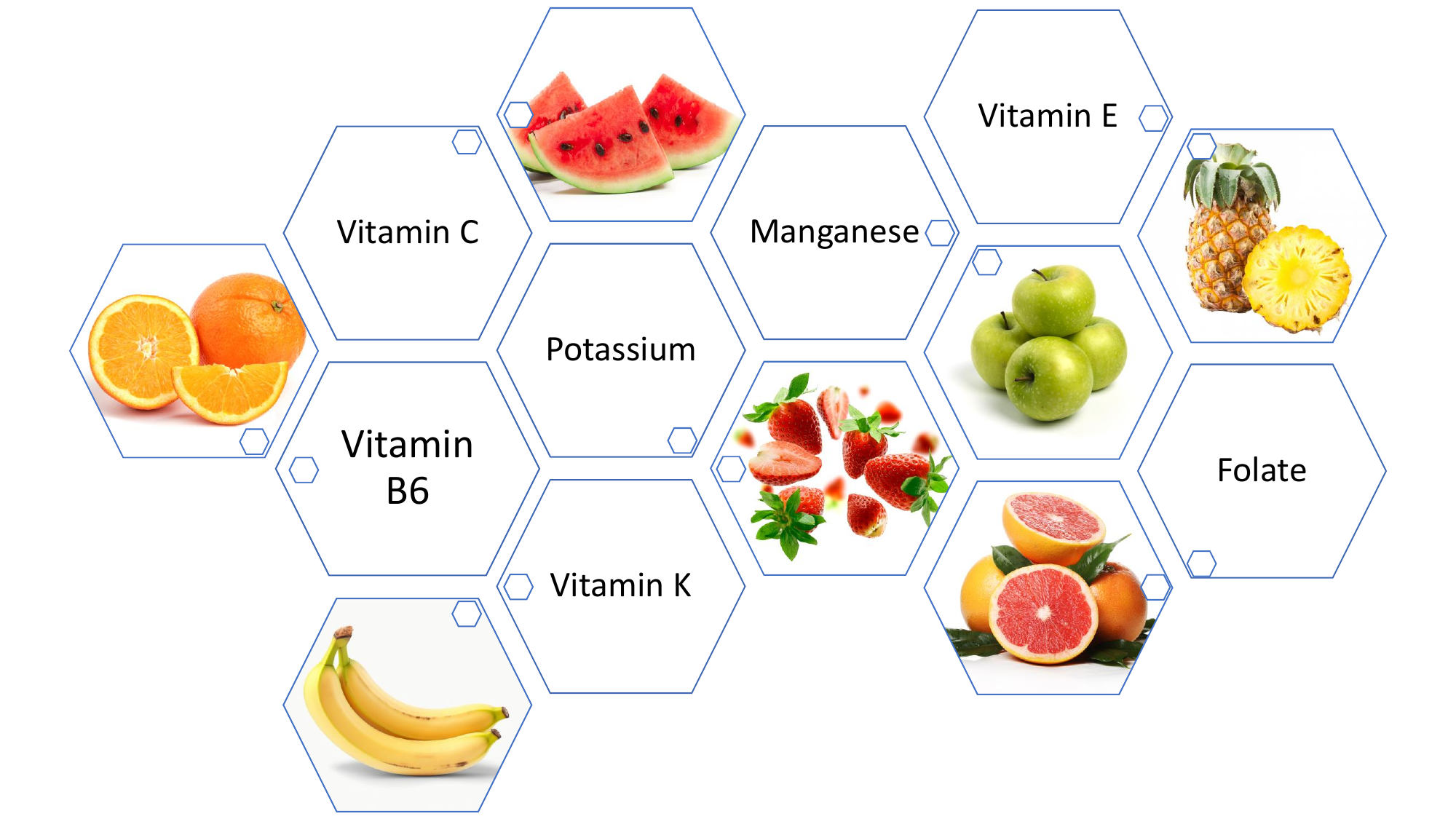}}
\caption{The vitamins FruitPAL 2.0 promotes}
\label{fig:fruits-gruope}
\end{figure}

Fruits are considered healthier options for meals, as they come along with attached advantages concerning their health benefits. Fresh fruits generally are a good source of several key nutrients containing vitamins and dietary fiber that have specific considerable human health benefits \cite{FruitPAL2}. Fig \ref{fig:fruits-gruope} presents an illustration of the many vitamins that can be obtained from eating fruit \cite{FruitPAL2}. They do not show any side effects when being consumed in large amounts; therefore, fruits should be consumed during every meal. Because fruits are easily digestible and sweet, they are taken as meals or snacks. Many researches suggest taking fruit as part of the daily routine diet due to its excellent nutritious value \cite{HealthyFruits,SugarFruits,VegetablesandFruits}.

\begin{figure}[htbp]
\centerline{\includegraphics[width=0.7\textwidth]{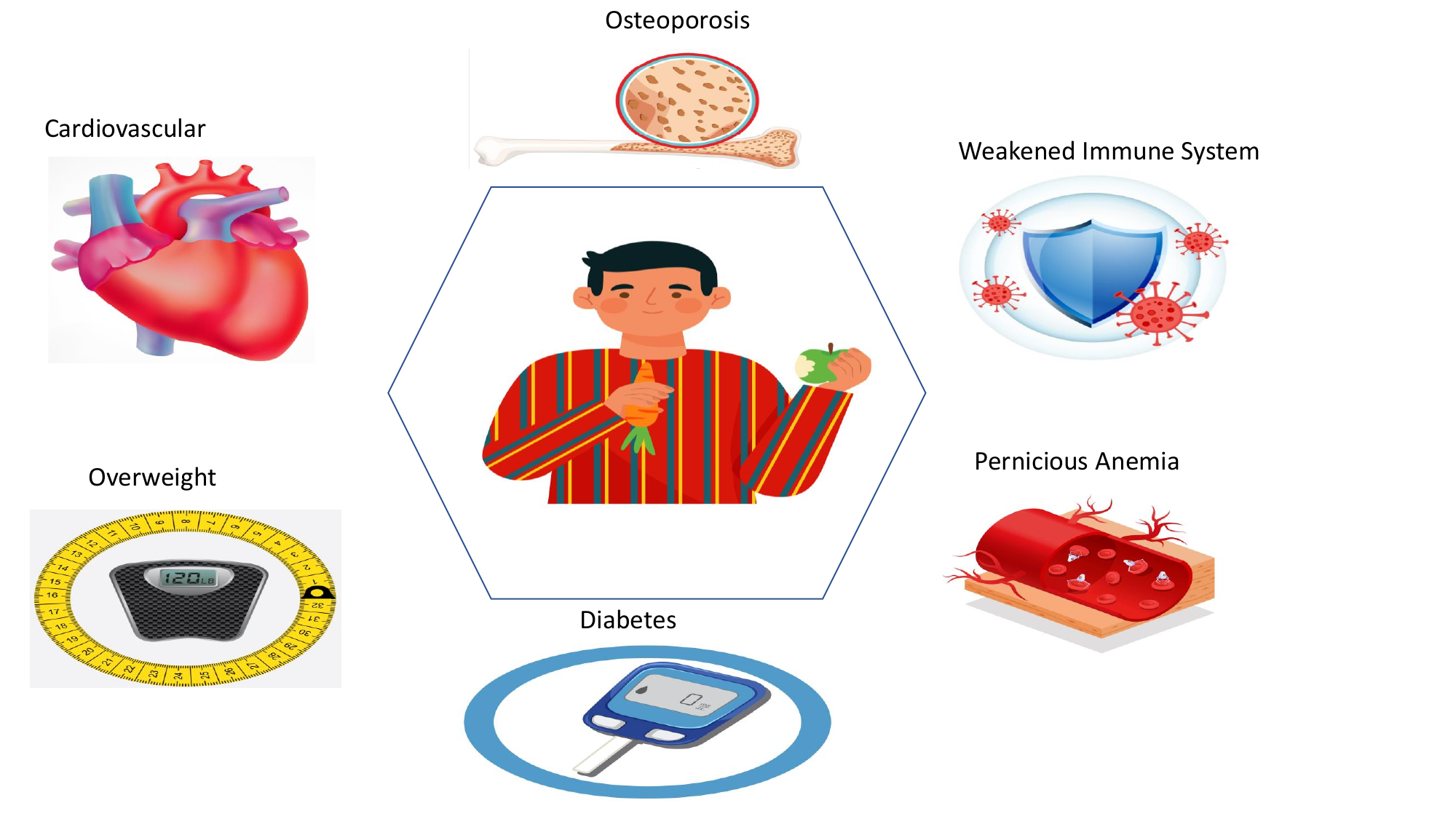}}
\caption{Preventable Disease by Eating Fruit.}
\label{fig:Diseases}
\end{figure}

Consuming fruits boosts the immune system due to their high vitamin content. Research shows that eating fruits helps lower the risk of diseases like cardiovascular conditions and metabolic syndrome \cite{Soluble}. Additionally, fruit intake is linked to a decrease in the consumption of unhealthy and processed foods. A study found that having at least four to five servings of fruit daily can improve mood \cite{30Healthiest}. Regular fruit consumption also helps prevent vitamin deficiencies. Fig \ref{fig:Diseases} highlights diseases that can be prevented through fruit consumption.

Fruits are generally known as a health role of dietary component. However, in certain cases, fruit consumption can lead to potential negative impacts that may be significant on human health. People who suffer from an allergy to certain types of fruits should avoid those types. Because of the nutritional value of fruit as a food source, fruit allergy is common diseases that attack immunity \cite{FruitPAL}. In addition, people who suffer from diabetes or pre-diabetes should be very cautious when selecting fruits for consumption and focus mostly on the fruits containing low quantities of sugar\cite{SugarContent}. Fruits are considered healthy dietary sources except in cases when they cause sickness for the human body. In this paper, we present a detailed work to develop a healthcare cyber-physical system (H-CPS). The initial H-CPS is called FruitPAL, whereas the improved one is defined as FruitPAL 2.0.


FruitPAL, the real-time device, aims at the detection of fruits that cause allergies and warning the caregivers of people suffering from fruit allergens due to lack of recognition. The proposed device detects fruit allergens in the shortest time to protect the immune system by warning the caregivers. FruitPAL is designed to protect the lives of individuals with fruit allergies, contributing to advancements in smart healthcare.

FruitPAL 2.0, an updated version of FruitPAL, is a novel method designed to encourage consumers to incorporate fruit into their daily diet. The many health perks of fruit can be achieved with FruitPAL 2.0. The primary aim of monitoring the fruit intake of FruitPAL 2.0 is to provide valuable insights into  how fruit contains health nutrients.

\section{CONTRIBUTIONS OF THE CURRENT PAPER}
In this section, the novelties of FruitPAL and it's advanced version have been discussed. 

\paragraph{FruitPAL: } 

\begin{itemize}
     \item The technology automatically identifies fruits that can cause allergic reactions to the consumers.

	 \item It is characterized by its real-time nature, enabling it to yield results promptly.

     \item  An alert is automatically created, emailed, and delivered to the caregivers who have been assigned to that position up to the point when the system identifies a change in status. 
     
     \item Users have the ability to connect directly with the system through the utilization of the Global System for Mobile Communications (GSM), which provides a channel for such interaction.
     
\end{itemize}



\paragraph{FruitPAL 2.0: }

\begin{itemize}
     \item The proposed work automatically detects and analyzes the fruit that was eaten.

     \item FruitPAL 2.0 has ability to classify fifteen types of fruit.

     \item The nutrients information message is automatically sent to consumer on time.

	 \item The system uses Global System for Mobile Communications (GSM) to send the message to user.

\end{itemize}

\paragraph{FruitPAL vs FruitPAL 2.0:}
FruitPAL 2.0 is an upgraded version of FruitPAL. Its goal was to enable people to be motivated to eat fruits. In contrast, FruitPAL deals with the automatic detection of allergenic fruits. Both are based on the Allergic-fruit dataset \cite{Allergic-fruit}. While FruitPAL 2.0 shows improved metrics compared to FruitPAL due to the use of YOLOv5 V6.0, FruitPAL offers a faster response time due to the YOLOv8 model.

\section{RELATED RESEARCH}

The referenced research employs several methodologies to assess food quality. The Dietary Intake Monitoring System proposed by \cite{Ofei2014TheDI} measures food temperature and monitors changes in plate weight for patients. Stored fruit quality can be assessed using laser diagnostics \cite{6421173}, while baby food quality is evaluated with an electronic tongue to determine if it meets required standards \cite{1604076}. An application using a CNN model and the Fruits 360 dataset to classify fruits and display allergen information is mentioned in \cite{FrameworktoIdentifyAllergen}. Additionally, image segmentation was discussed in \cite{apple} to classify the quality of Manalagi apples in smart agriculture. While consuming high-quality fruit is valuable, there is still hesitation regarding its consumption. However, avoiding fruit can be beneficial if it triggers allergies.

The detection of food consumed can be achieved through the classification of sound. Monitoring food consumption using two microphones to capture sounds of swallowing and eating was mentioned on \cite{6038810}. Wearable device on the neck that can detect sound of solid and liquid food and keeps track eating habits was referenced on \cite{6855620}. A wearable sensor Level with a microphone and a camera is meant to identify chewing activity based on sound features and video sequence analysis, recording both the order in to show the consumption rate \cite{6200559}. Nevertheless, the analysis of sound can be inefficient and lead to the dissemination of inaccurate information.


Food consumption detection can be achieved through sound classification. Monitoring food intake by using two microphones to capture swallowing and eating sounds was discussed in \cite{6038810}. A neck-worn device that can detect sounds from solid and liquid foods and track eating habits was referenced in \cite{6855620}. Another wearable sensor, Level, equipped with a microphone and camera, is designed to identify chewing activity based on sound features and video analysis, recording the sequence to display the consumption rate \cite{6200559}. However, sound-based analysis can be inefficient and may lead to inaccurate information dissemination.

Publications on Smart Agriculture range from counting the number of crops in a specific area. Object detection in greenhouses to count and identify pepper plants is involved \cite{Automaticfruit}. Faster RCNN is applied on robotic to detect the fruit \cite{Afruitdetection}. Also ResNet, object detection with ImageNet dataset is used on robotic to count the fruit and flowers \cite{Rahnemoonfar}. A computerized robotic detecting mangoes, almonds, and apples in orchards was discussed in \cite{detectioninorchards}. The research works mentioned above count fruits that are planted in crops and assist the substance of the food chain.

There are two types of consequences when any fruit is consumed; one is beneficial to human bodies, and another one may also have some adverse effects within the body. The proposed devices will debate on the consequences of human health in both positive and negative directions. Table \ref{table:priorworks} presents many models and datasets that have been used in related research. Notably, FruitPAL and FruitPAL 2.0 are completely automated devices containing efficiency models; YOLOv8 and YOLOv5 V6.0, respectively.

\begin{scriptsize}
\begin{table}[htbp]
  \caption{Details of the models and datasets used in the existing works}
  \centering
  \begin{tabular}{|l|l|l|l|}
    \hline
    \textbf{Works} & \textbf{Model} & \textbf{Dataset} \\
    \hline
    \hline

       B. Rohini \cite{FrameworktoIdentifyAllergen}  & Image classification (CNN) & Fruits 360 \\ 
    \hline
         M. Muladi \cite{apple}  & Image classification (Backpropagation) & Custom Dataset \\ 
    \hline
         S. Päßler \cite{6038810}  & Support Vector Machine &Custom Dataset \\ 
    \hline
         H. Kalantarian \cite{6855620}  & Vibration Sensor &Custom Dataset\\ 
    \hline
     J. Liu \cite{6200559}  & Extreme Learning Machines &custom dataset\\ 
    \hline 
       Y. Song \cite{Automaticfruit}  & Image classification (CNN) & PASCAL-VOC \\ 
    \hline
    Sa, Inkyu \cite{Afruitdetection} & Object detection ( Faster R-CNN) & ImageNe\\
    \hline
    M. Rahnemoonfar \cite{Rahnemoonfar} & Object detection (ResNet) & ImageNet\\
      \hline
      S. Bargoti \cite{detectioninorchards} & Object detection ( Faster R-CNN) & PASCAL-VOC\\ 
    \hline
      M. Alakananda \cite{iLog}& Object Detection (API) & Food-A-Pedia\\
    \hline   
     B.  Issam \cite{Handwritten}& Image classification (CNN) & CVL single digit\\
    \hline
     M. Hossain \cite{IndustrialApplications}& Image classification (CNN) & supermarket produce\\
    \hline
     \textbf{A. Alkinani} \cite{FruitPAL} 
     &  Object Detection (YOLOv8) & Allergic-fruit\\
     \hline
     \textbf{A. Alkinani} \cite{FruitPAL2} &  Object Detection (YOLOv5 V6.0) & Allergic-fruit\\
    \hline
  
  \end{tabular}
  \label{table:priorworks}
\end{table}
\end{scriptsize}
\bibliographystyle{unsrt}  

\section{PROPOSED HEALTHCARE CYBER-PHYSICAL SYSTEM}


\subsection{Overview}
A Cyber-Physical System (CPS) integrates the physical world with computational systems by collecting data via sensors or cameras and processing it locally or in the cloud. FruitPAL and FruitPAL 2.0, as CPS devices, have significantly impacted smart healthcare. FruitPAL helps caregivers monitor fruit allergies by detecting specific fruits that may trigger reactions and providing immediate alerts using the highly accurate YOLOv8 model. On the other hand, FruitPAL 2.0 encourages daily fruit consumption by identifying and analyzing 15 types of fruit and sending motivational messages to participants, promoting healthier dietary habits.


\subsection{FruitPAL}

\begin{figure*}[htbp]
\centerline{\includegraphics[width=0.8\textwidth]{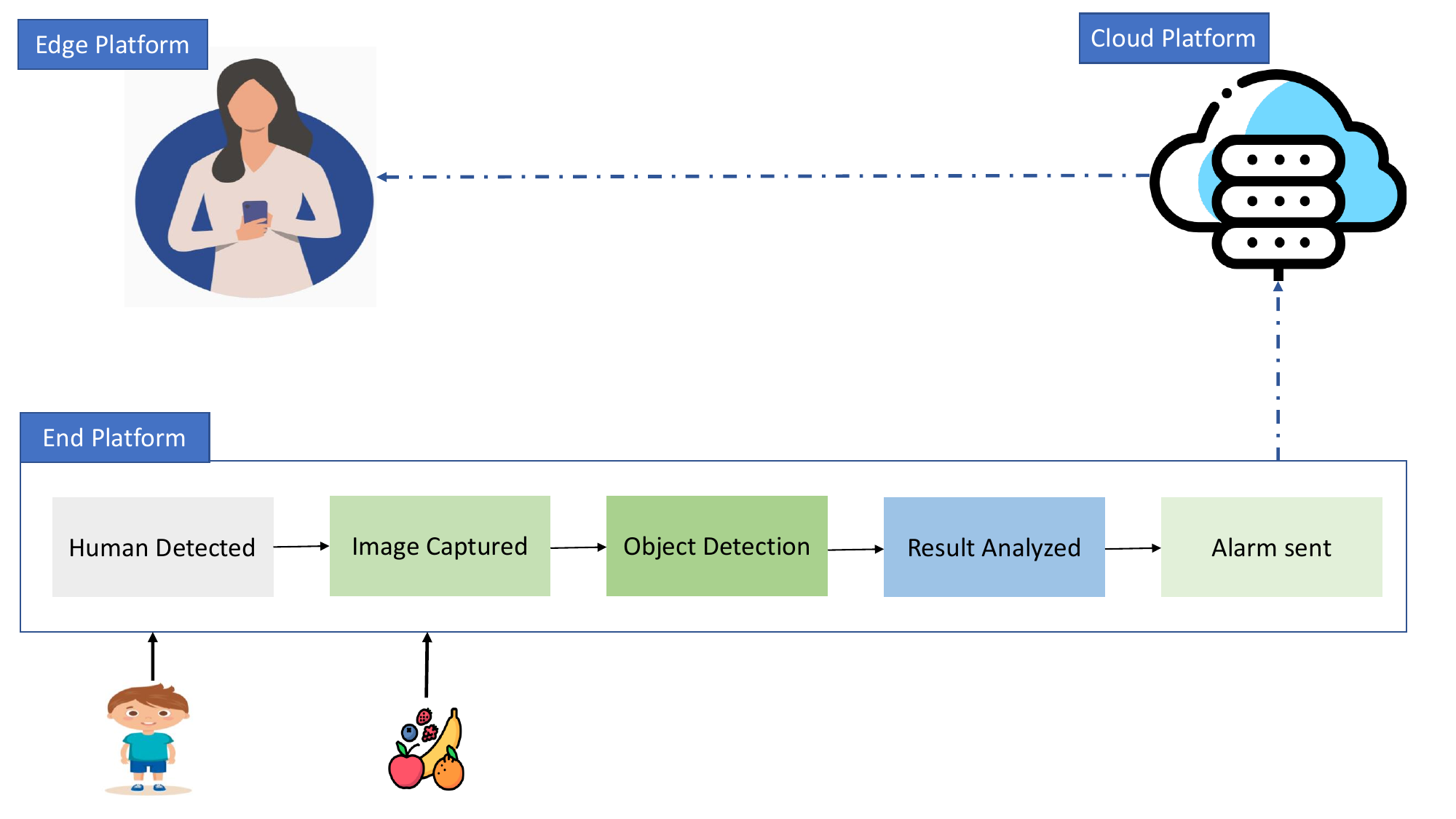}}
\caption{ FruitPAL Method}
\label{fig:FA Method}
\end{figure*}

The proposed FruitPAL method is illustrated in Fig \ref{fig:FA Method}. The system activates through the End Platform upon detecting human movement via the PIR sensor. Once the camera captures an image, the YOLOv8 algorithm identifies any fruit present. The object detection results are then analyzed, and if necessary, an alert is sent to the Cloud Platform. The cloud processes this alert and promptly relays it to the caregiver.

\subsubsection{Computing Platforms}
\begin{figure*}[htbp]
\centerline{\includegraphics[width=0.8\textwidth]{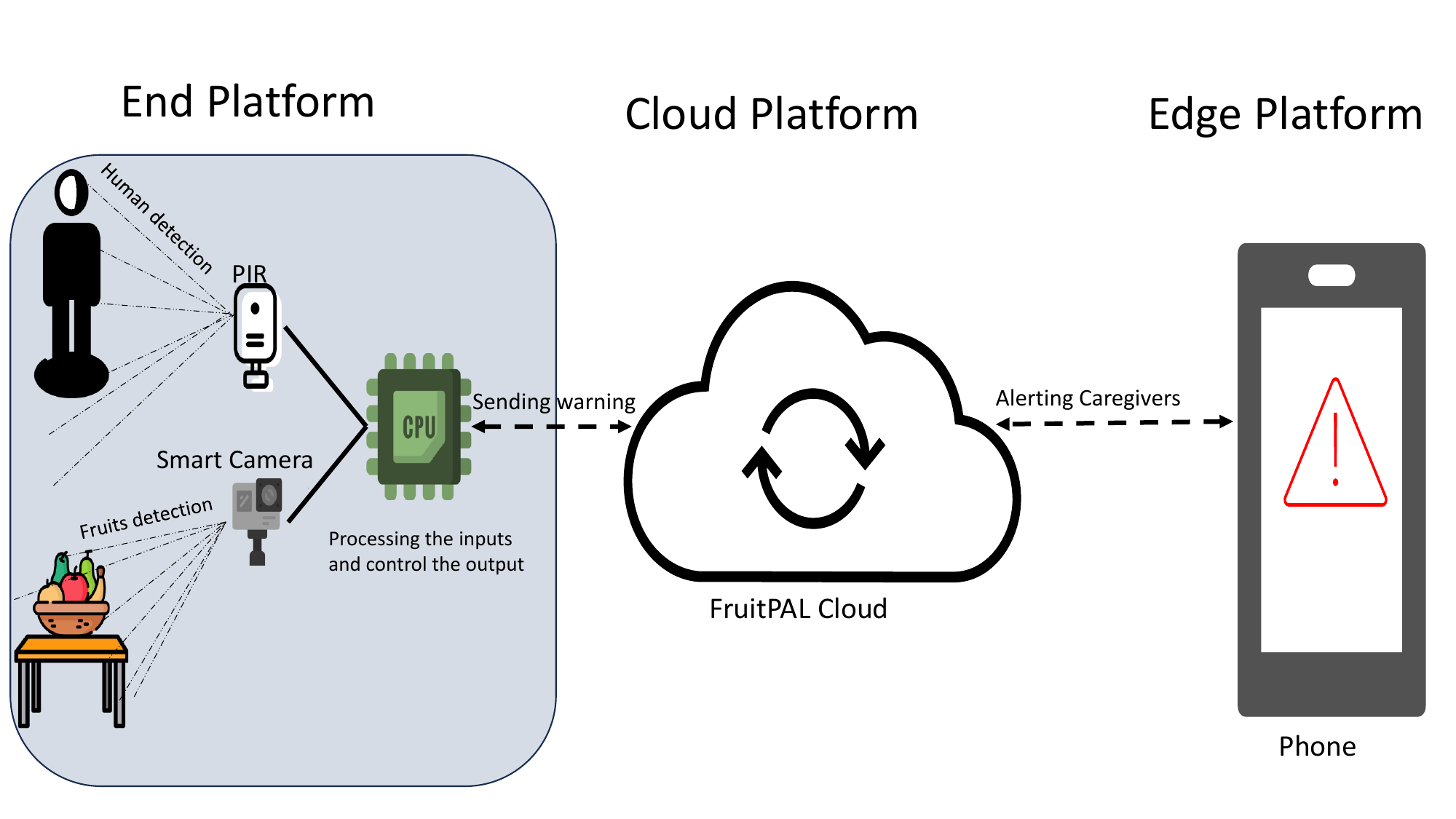}}
\caption{ Three Different Computing Platforms Visible in FruitPAL's Architecture: End Platform, Cloud Platform, and Edge Platform}
\label{fig:FA design}
\end{figure*}

In Fig \ref{fig:FA design} provides a system-level overview of the proposed FruitPAL concept. The platform integrates both a Passive Infrared (PIR) sensor and a smart camera at the same level, allowing for early and accurate hazard prediction. Once powered on, the camera captures a sequence of images, which are automatically processed and stored locally. Data is only sent to a notification platform when human movement is detected by the PIR sensor. Notifications are then sent to caregivers’ mobile devices, where caregivers can choose to dismiss them as needed.

The system is distributed across multiple computing platforms, all connected via wireless communication. The End Platform is responsible for image capture, motion detection, and initial image processing. The Cloud Platform acts as the communication hub for the system, linking its various components. Finally, the Edge Platform receives notifications from the cloud and enables the caregiver’s response to be sent back to the system.

\begin{figure}[htbp]
\centerline{\includegraphics[width=0.8\textwidth]{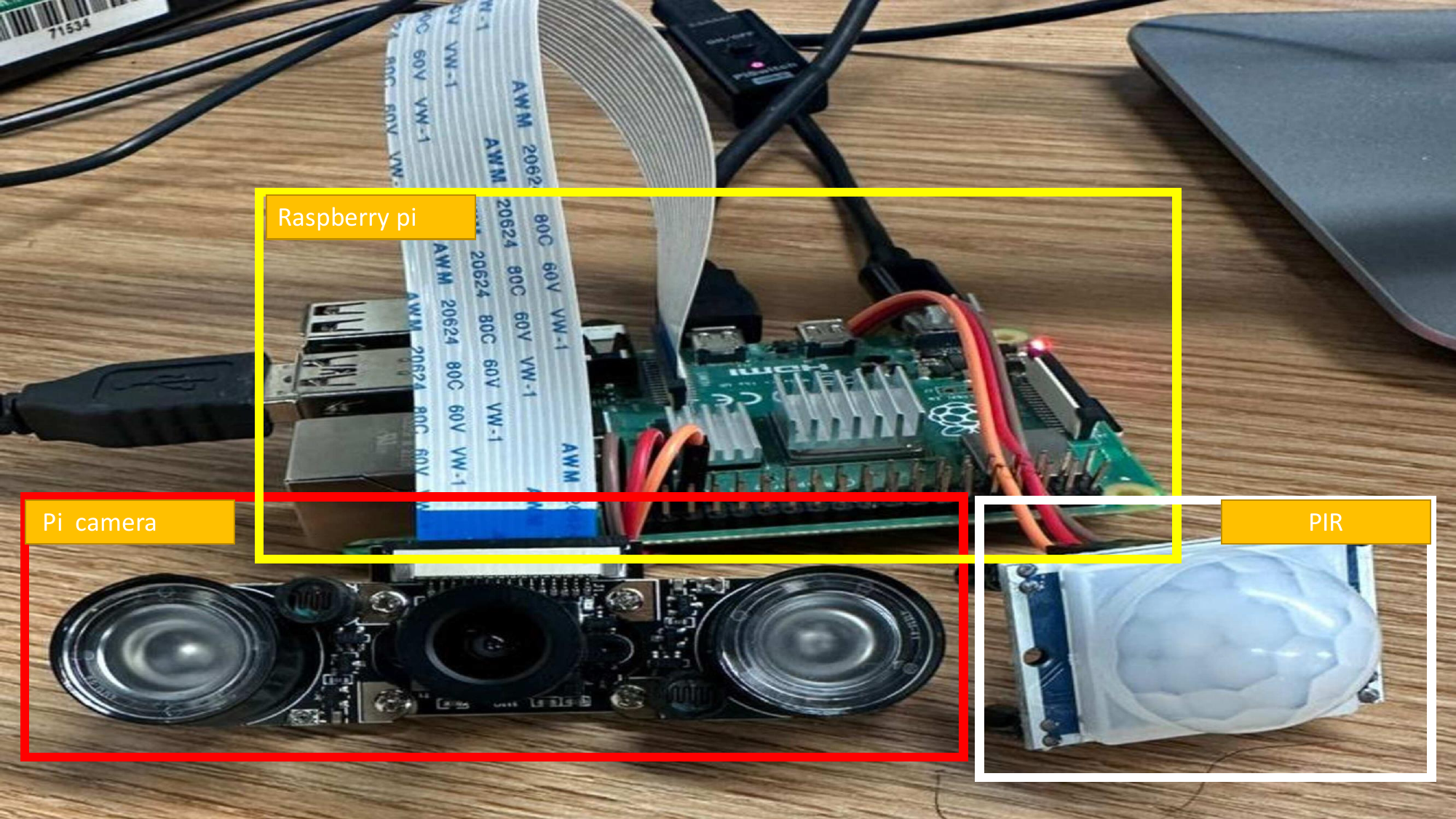}}
\caption{End Platform in FruitPAL}
\label{fig:END PLATFORM}
\end{figure}

\paragraph{End Platform:}
At this stage, FruitPAL identifies allergy-causing fruits and sends the gathered data to the next phase. In Fig \ref{fig:END PLATFORM} illustrates the end platform components: the PIR sensor and smart camera. These components work in sync to enable a rapid response. The PIR sensor detects motion from humans and animals based on body temperature \cite{PIR} and should be positioned in a broad area for optimal motion detection of the individual with allergies.

The second key component, the smart camera, detects allergenic fruits. High-quality visual capabilities are essential to ensure accuracy in detection results.

\paragraph{Cloud Platform \& Edge Platform:}
The Cloud Platform serves as a bridge between the End Platform and the Edge Platform, with communication facilitated by a Wi-Fi connection. Upon receiving alerts from the End Platform, the Cloud Platform notifies caregivers through the Edge Platform. To confirm the presence of a potential allergen, caregivers must respond to the alert. The warning reaches the caregiver as a phone call with a voice message indicating, "Allergen detected – danger present." The alert ceases once the individual with allergies moves away or the caregiver takes appropriate action. FruitPAL is vital in safeguarding lives.

\subsubsection{Training Protocol}
FruitPAL is a real-time system designed to accurately predict the risk of fruit allergens. Various algorithms are available for real-time object detection, including Faster-RCNN, DPM, and the YOLO family. The dataset \ref{Dataset cp} for FruitPAL was trained using YOLOv8, a model from the YOLO family, chosen for its high accuracy and fast object detection capabilities. The YOLO family has made a notable impact in object detection due to its single-stage architecture \cite{IntroductiontotheYOLOFamily}. Unlike other algorithms, the YOLO family does not require a Region Proposal Network (RPN) \cite{Understanding}. The training was performed using the YOLOv8s model, where "s" denotes a small-weight configuration with 11.2 million parameters, making it suitable for deployment on edge devices \cite{WhatisYOLOv8}.

\subsubsection{Experiment} \label{Yolov8m}

\begin{figure}[htbp]
\centerline{\includegraphics[width=15cm,height=45cm,keepaspectratio]{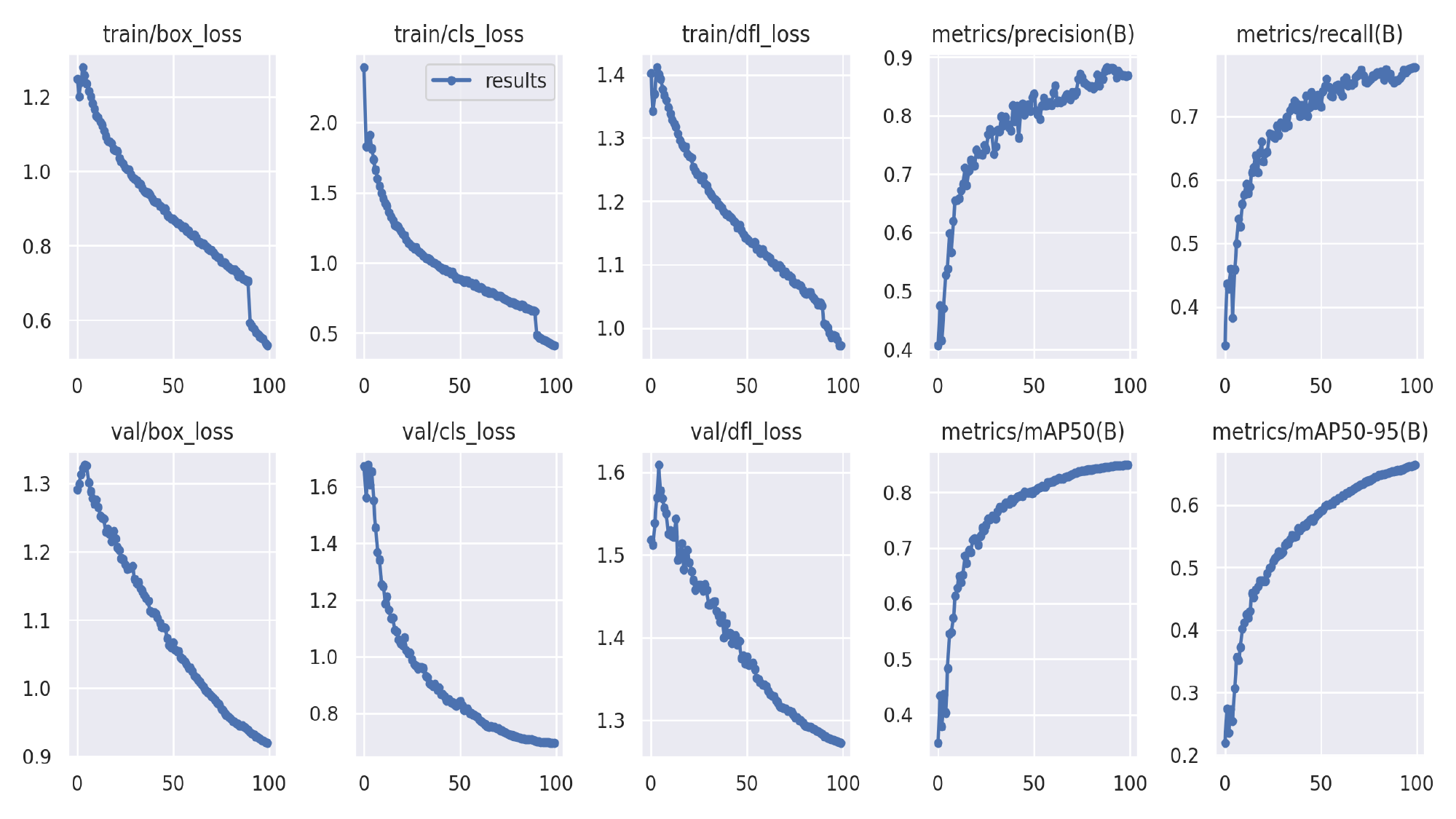}}
\caption{FruitPAL Metrics Model}
\label{fig:MetricEvalaution}
\end{figure}

\begin{figure}[htbp]
\centerline{\includegraphics[width=15cm,height=15cm,keepaspectratio]{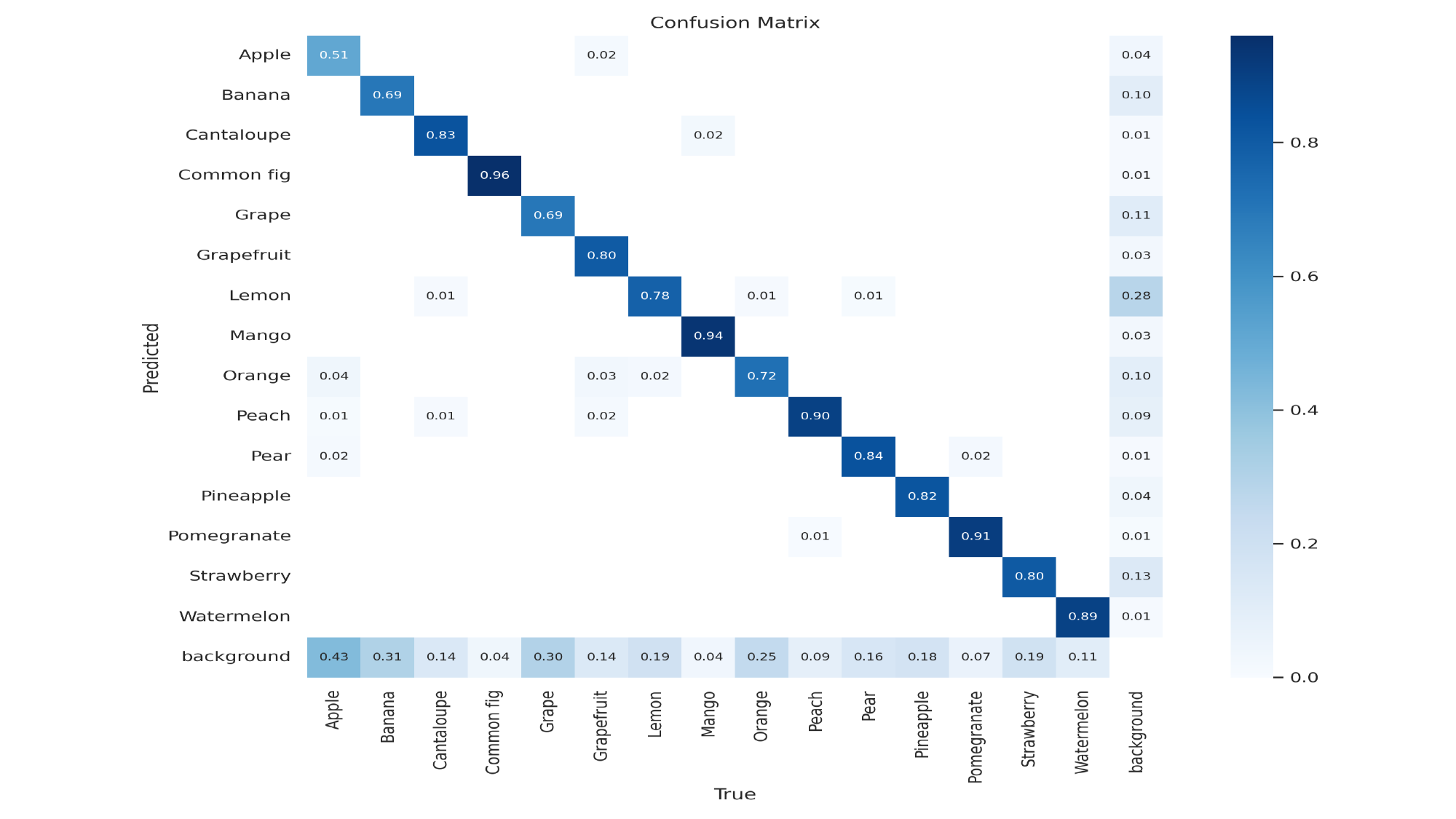}}
\caption{FruitPAL Confusion Matrix}
\label{fig:confusion-matrix}
\end{figure}

FruitPAL is a portable electronic device that can be positioned flexibly in any location. The model’s computational performance was evaluated on a system with an Intel Core i7-7700 processor, clocked at 3.60 GHz, and 16 GB of RAM. For implementation, a Raspberry Pi 4 with 4GB RAM was used in the end platform, as shown in Fig \ref{fig:END PLATFORM}. Model training was conducted on an A100 GPU (a Google Colab GPU) with 40GB of RAM, completing 100 epochs in approximately 3 hours. In Fig \ref{fig:MetricEvalaution} presents the model evaluation, detailing training metrics—such as box loss, class loss, and DFL loss—as well as validation metrics, including box loss, class loss, and DFL loss. Key performance indicators like precision, recall, and mAP50-95 are also evaluated in this study. Table \ref{table:formatting} presents the summary of the metrics. The confusion matrix In Fig \ref{fig:confusion-matrix} presents the synopsis performance for each class of the model.

\begin{figure}[htbp]
\centerline{\includegraphics[width=14cm,height=25cm,keepaspectratio]{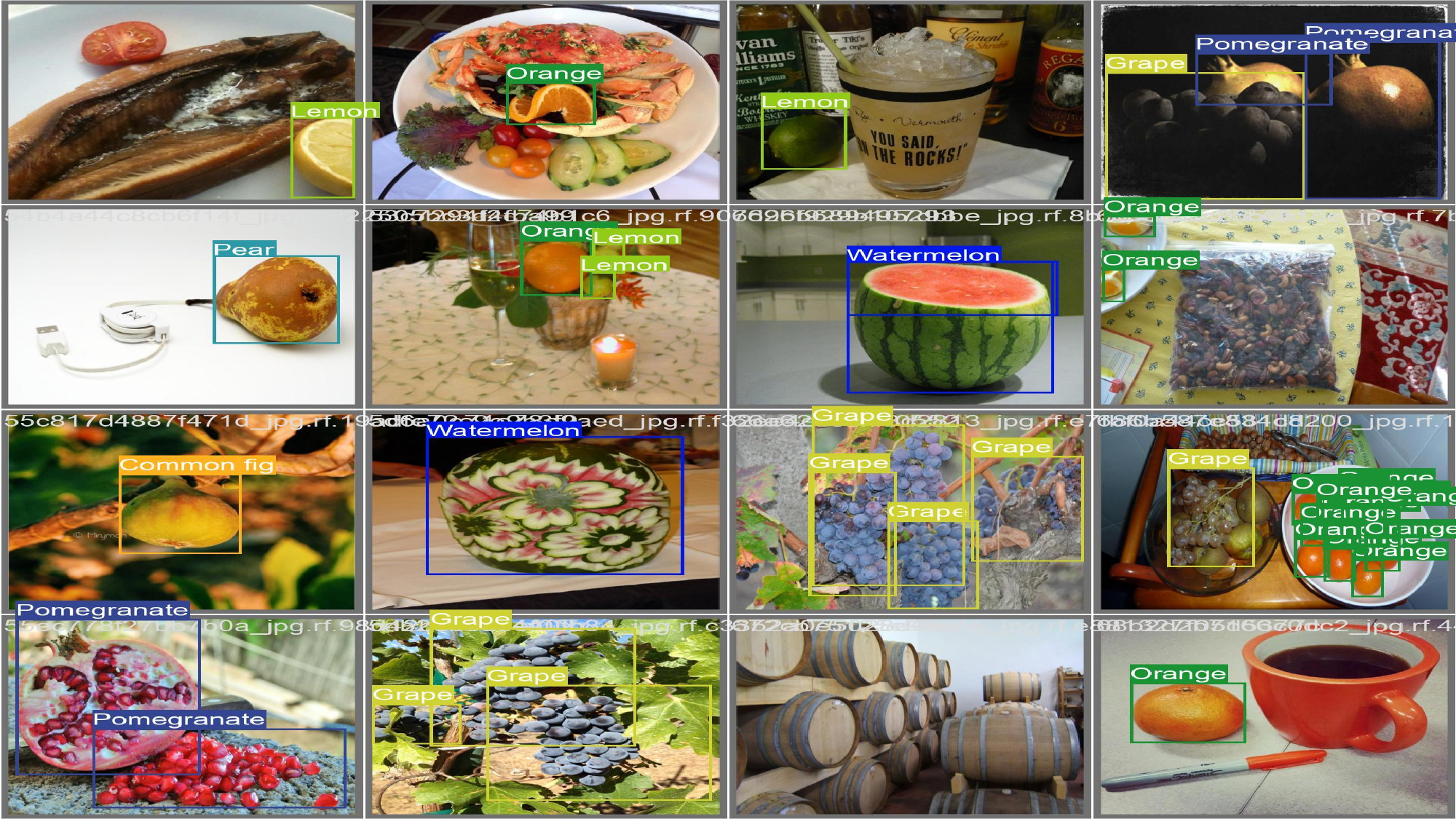}}
\caption{Object detection by FruitPAL}
\label{fig:allegic fruit}
\end{figure}

In Fig \ref{fig:allegic fruit}, FruitPAL is shown accurately detecting fruits among other objects, highlighting its ability to identify potential dangers precisely. To enable seamless integration with the FruitPAL device, the YOLOv8 model was converted to TensorFlow Lite using ONNX. This integration allows individuals with allergies to avoid life-threatening fruits effectively

\begin{scriptsize}
\begin{table}[htbp]
 \caption{FruitPAL Metrics}
  \centering
  \begin{tabular}{|l|l|}
    \hline
    \textbf{Metrics} & \textbf{Value}\\
    \hline
    \hline
    Precision & 86\%  \\
    \hline
    Recall & 77\% \\
    \hline
     mAP50 & 84\% \\
    \hline
    mAP50-95 & 66\%\\
    \hline 
    Parameters & 25m\\
    \hline
  \end{tabular}
  \label{table:formatting}
 
\end{table}
\end{scriptsize}

\subsubsection{Hardware for Prototyping}
The hardware components within the FruitPAL system exhibit high levels of effectiveness and efficiency. Nevertheless, it is possible to modify the PIR sensitivity as required. The calculation of the sensitivity of PIR sensor is imperative in order to effectively detect human motion \cite{PIR}. The sensitivity of PIR can be adjust using the equation below.
:
\begin{quote}

Ti = 24 * R9 * C7.

Ti is a sensitivity

Playing with the values of resistor (R9) or capacitor (C7) change of the value of the sensitivity.   
\end{quote}
The higher and lower limits of responsiveness can be established based on the given equation.
\begin{quote}
The highest point of responsiveness:
Ti = 24 * 1m * 0.01uF = ~ 2.4 seconds

The lowest point of responsiveness:

TX = 24 * 1m * 0.05uF = 12 seconds
   
\end{quote}



\subsection{FruitPAL 2.0}

\begin{figure}[htbp]
\centerline{\includegraphics[width=15cm,height=50cm,keepaspectratio]{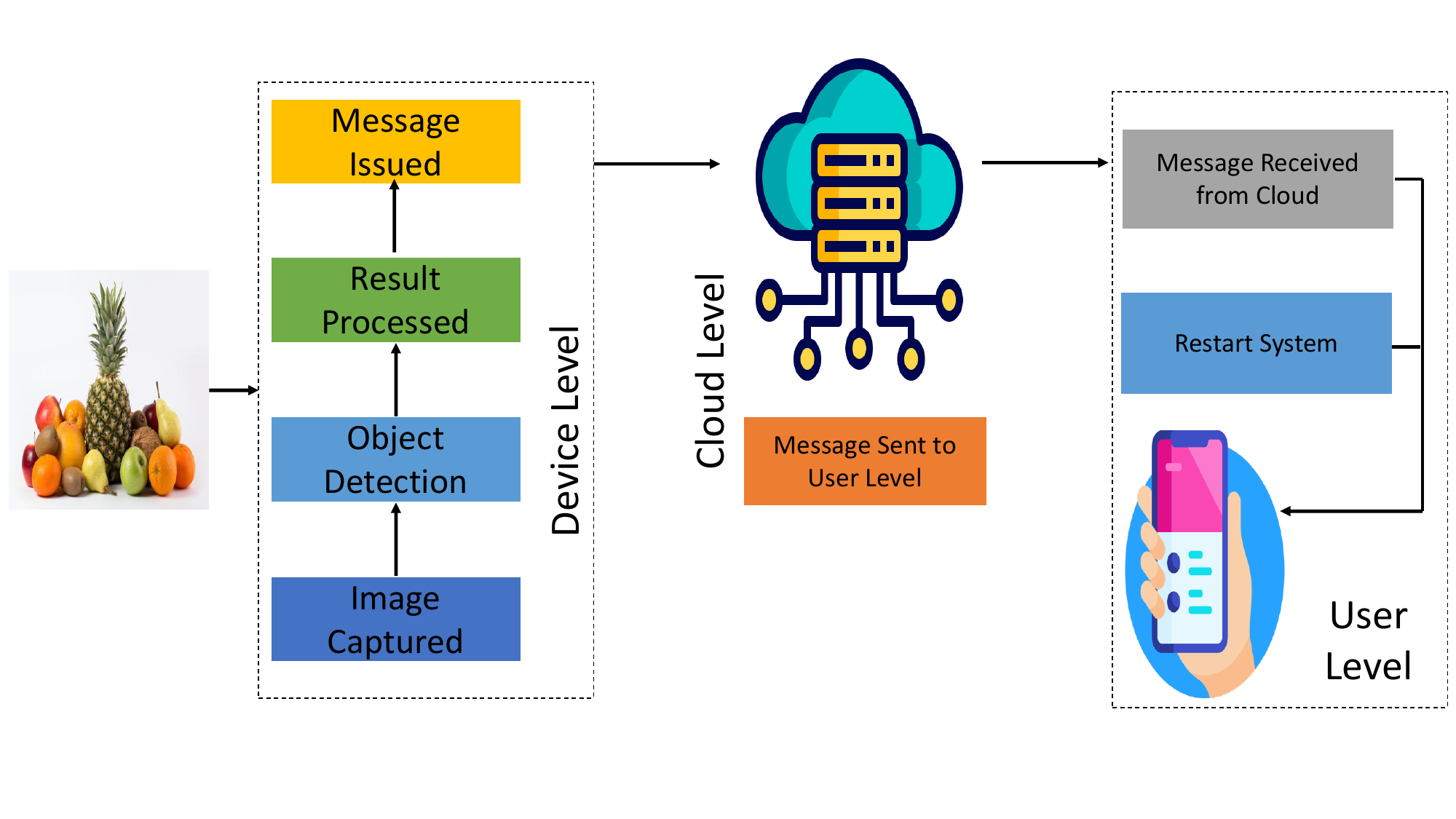}}
\caption{Workflow for Levels in FruitPAL 2.0}
\label{fig:FruitPAL 2.0}
\end{figure}

In Fig \ref{fig:FruitPAL 2.0} illustrates how each Level works to identify the fruit being consumed and deliver a text message. All functions are automatically full. Capturing images, detecting image, and analyzing the results is done on Device Level. Once the message is issued, the message is sent to User Level by Could Level. The message is sent everyday at specific time to avoid annoying the users. Restart the system can be done by the User Level when new fruits are added to the plate. The users may get a boost in motivation by receiving a daily message.

\subsubsection{Computing Platforms}
\begin{figure}[htbp]
\centerline{\includegraphics[width=25cm,height=25cm,keepaspectratio]{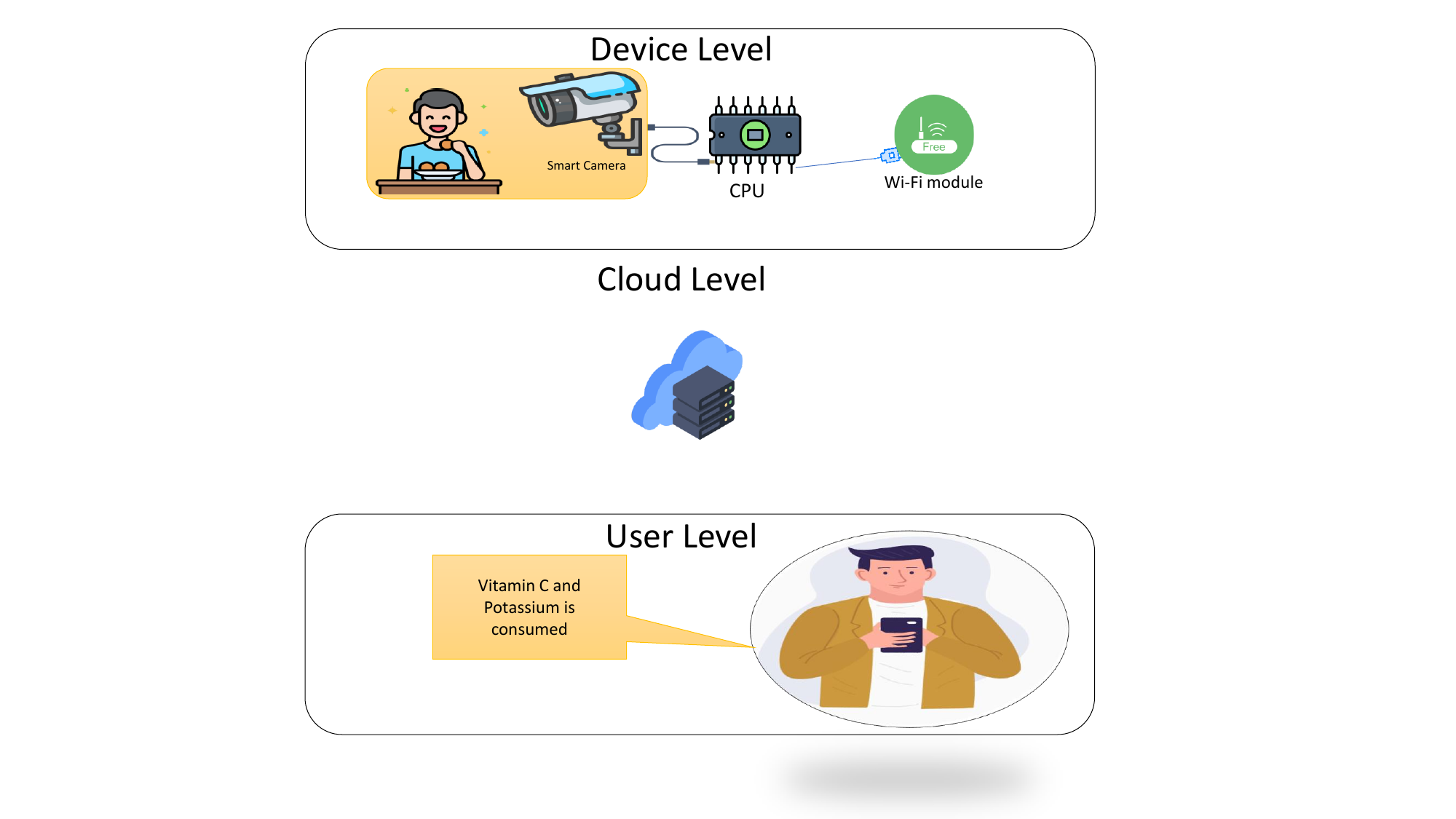}}
\caption{ FruitPAL 2.0's architecture shows three different computing Levels: Device Level, Cloud Level, and User Level}
\label{fig:FA2 design}
\end{figure}

Three levels are illustrated in Fig \ref{fig:FA2 design} to explain FruitPAL 2.0's architecture. The camera captures a photo when the power supply is turned on. The photo will be detected by the YOLOv5m V6.0 model. The results can be analyzed and sent to the next level, which is Cloud Level. The feature of the Could Level is to send a text message to phone on the User Level at a specific time. The text message contains a list of all the vitamins in the fruit that was eaten. Wireless commutation is applied between Levels. The majority of work will be done by Device Level to reduce the cloud cost. The Cloud Level is a commutation station between the Device Level and User Level.

\paragraph{Device Level:}
The image is taken by smart camera once the device is active. YOLOv5m V6.0 mode is applied on captured image. The initial fruit list will be stored after the fruit has been identified. Every hour, the fruit in the dish will be photographed and detected. New list will be created and Compared with the original list to show the consumed fruits. The nutrient value from each eaten fruit will be provide based in Table \ref{table:Value2}. The consumed nutrients will be written on a text message, and it will be provided in a specific time. In Table \ref{table:Value2} shows some of nutritional for each fruit groups\cite{HealthyFruits}. The message provides a summary of the nutritional that was consumed throughout the day. However, the system automatically restarts every morning to update the original list.

\begin{scriptsize}
\begin{table*}[htbp]
 \caption{Fruits Nutritional}
  \centering
  \begin{tabular}{|l|l|l|}
    \hline
    \textbf{Group} & \textbf{Value}&\textbf{Fruit}\\
    \hline
    \hline
    Citrus Fruits & Vitamin C and Potassium & Grapefruit, Lemon, and Orange.\\
    \hline
    Tropical Fruits& Vitamin B6 and C  & Banana, Mango, and Pineapple.\\
    \hline
    Pome Fruits & vitamin C and Manganese & Apple, Common fig, and Pomegranate.\\
    \hline
    Stone Fruits & vitamins A, C, and E  & Peach and Pear.\\
    \hline
    Melons Fruits & Vitamins A and C  & Cantaloupe and Watermelon.\\
    \hline
    Berries Fruits & Vitamin K and Folate  & Grape and Strawberry.\\
    \hline

  \end{tabular}
  \label{table:Value2}
 
\end{table*}
\end{scriptsize}

\begin{figure}[htbp]
\centerline{\includegraphics[width=10cm,height=25cm,keepaspectratio]{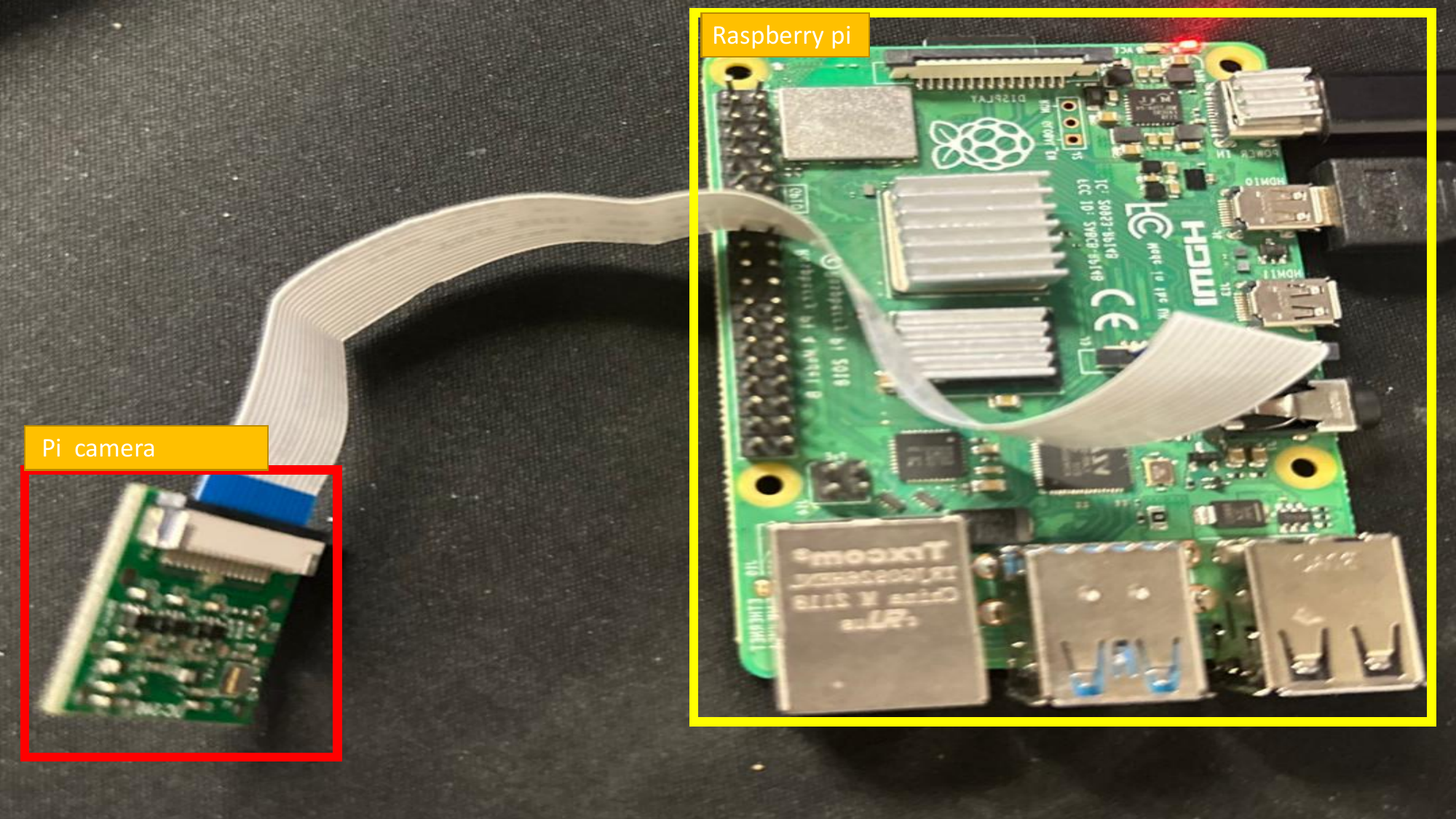}}
\caption{Device Level in FruitPAL 2.0}
\label{fig:Device Level}
\end{figure}

\paragraph{Cloud Level \& User Level: }
The link between Device Level and User Level is Cloud Level. The text message is created by Device Level. It is sent to User Level by Cloud Level. Users do not need Internet connection to receive the message duo to GSM.



\subsubsection{Training Protocol} \label{Yolov5m}

\begin{figure}[htbp]
\centerline{\includegraphics[width=14cm,height=25cm,keepaspectratio]{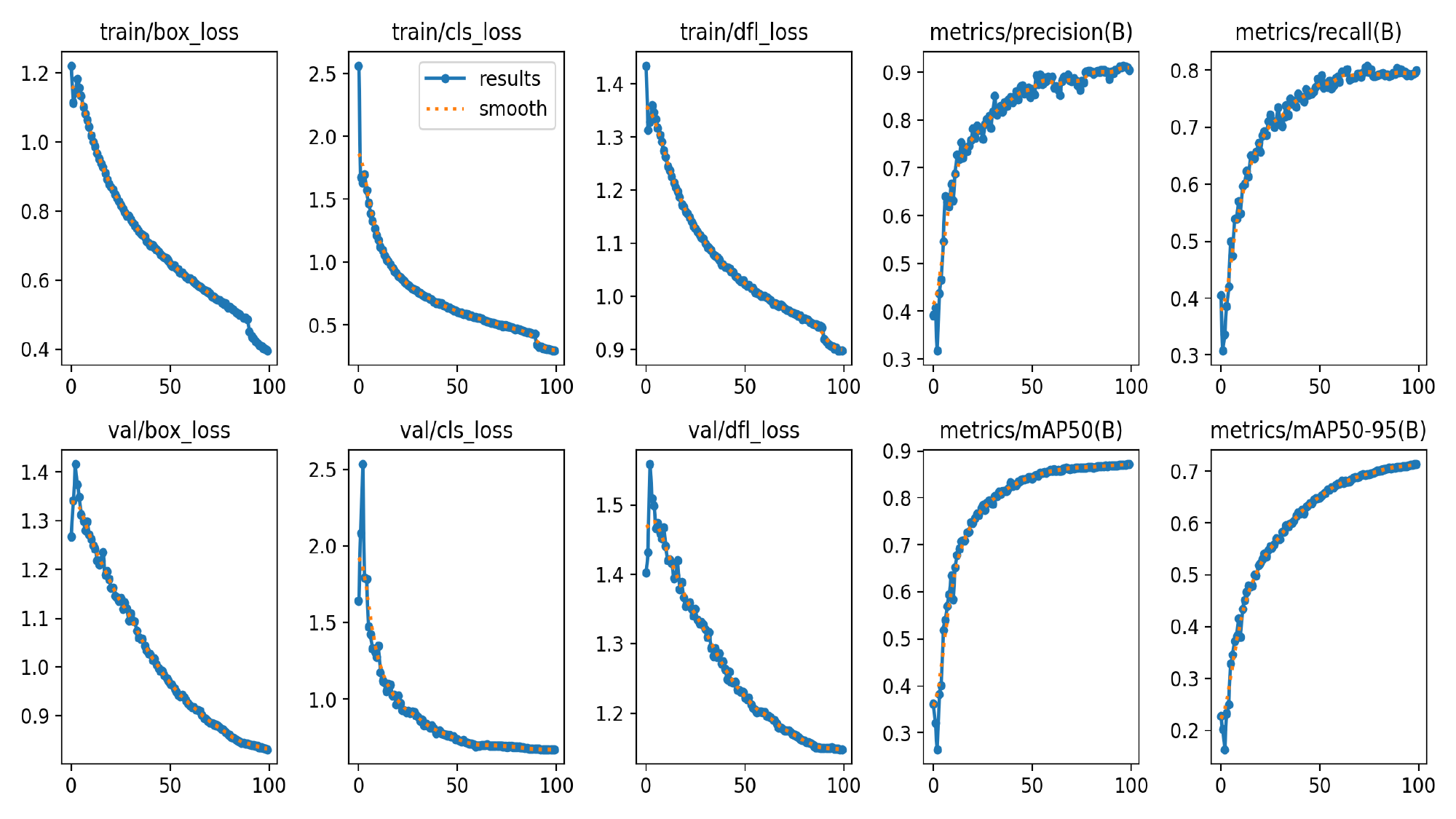}}
\caption{FruitPAL 2.0 Model Evaluation}
\label{fig:evaluate2}
\end{figure}

\begin{figure}[htbp]
\centerline{\includegraphics[width=14cm,height=25cm,keepaspectratio]{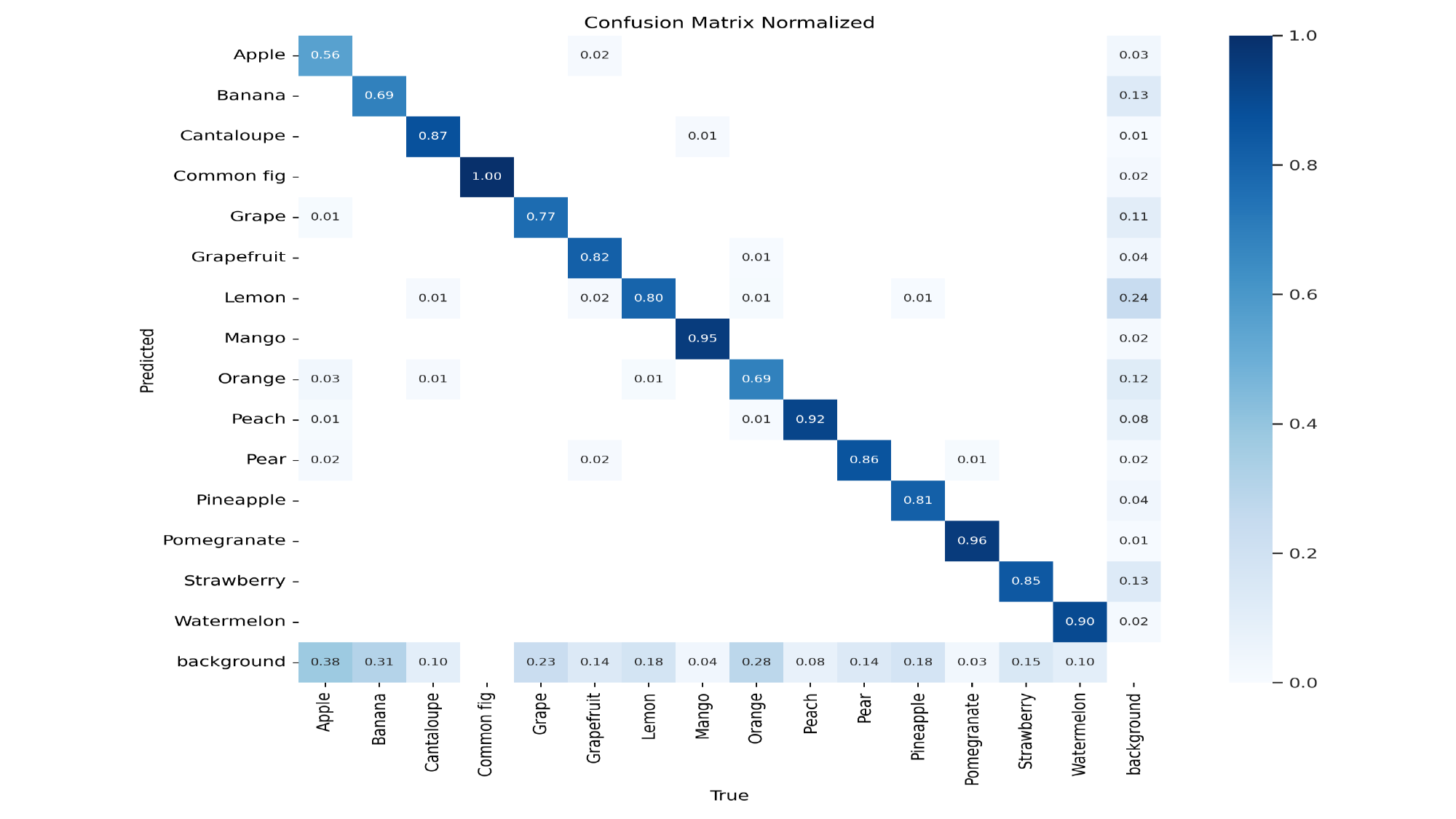}}
\caption{FruitPAL 2.0 Confusion Matrix}
\label{fig:confusion-matrix2}
\end{figure}

The FruitPAL 2.0 is a mobile electronic device that should be placed in dinning room to monitor the fruit consumption. YOLOv5m V6.0 has been utilized as the object detector in ``FruitPAL 2.00''. High timer respond and accurate is achieved duo to high quality dataset \ref{Dataset cp}. The T4 GPU, a Google Colab GPU, was used for training the model, and it has 40GB of RAM. The model was evaluated on Microsoft Surface Pro 7 that has Intel I5-1035G4 CPU, 8GB RAM, and 256GB SSD. In Fig \ref{fig:evaluate2} illustrates the model evaluation process for the object detection model, specifically focusing on the analysis of train box loss, class loss, and DFL loss. likewise, the above-mentioned graphic serves to exemplify many forms of validation, such val box loss, class loss, and DFL loss. The metrics of precision, recall, mAP50, and mAP50-95 are displayed in Fig. \ref{table:formatting}. The performance for each class of the model is illustrated in Fig \ref{fig:confusion-matrix2}. Open Neural Network Exchange (ONNX) is involved to exchange Yolov5 model from PyTorch to TensorFlow lite. TensorFlow lite is suitable for end device \cite{Lite}.

\subsubsection{Experiment}
\begin{figure}[htbp]
\centerline{\includegraphics[width=14cm,height=25cm,keepaspectratio]{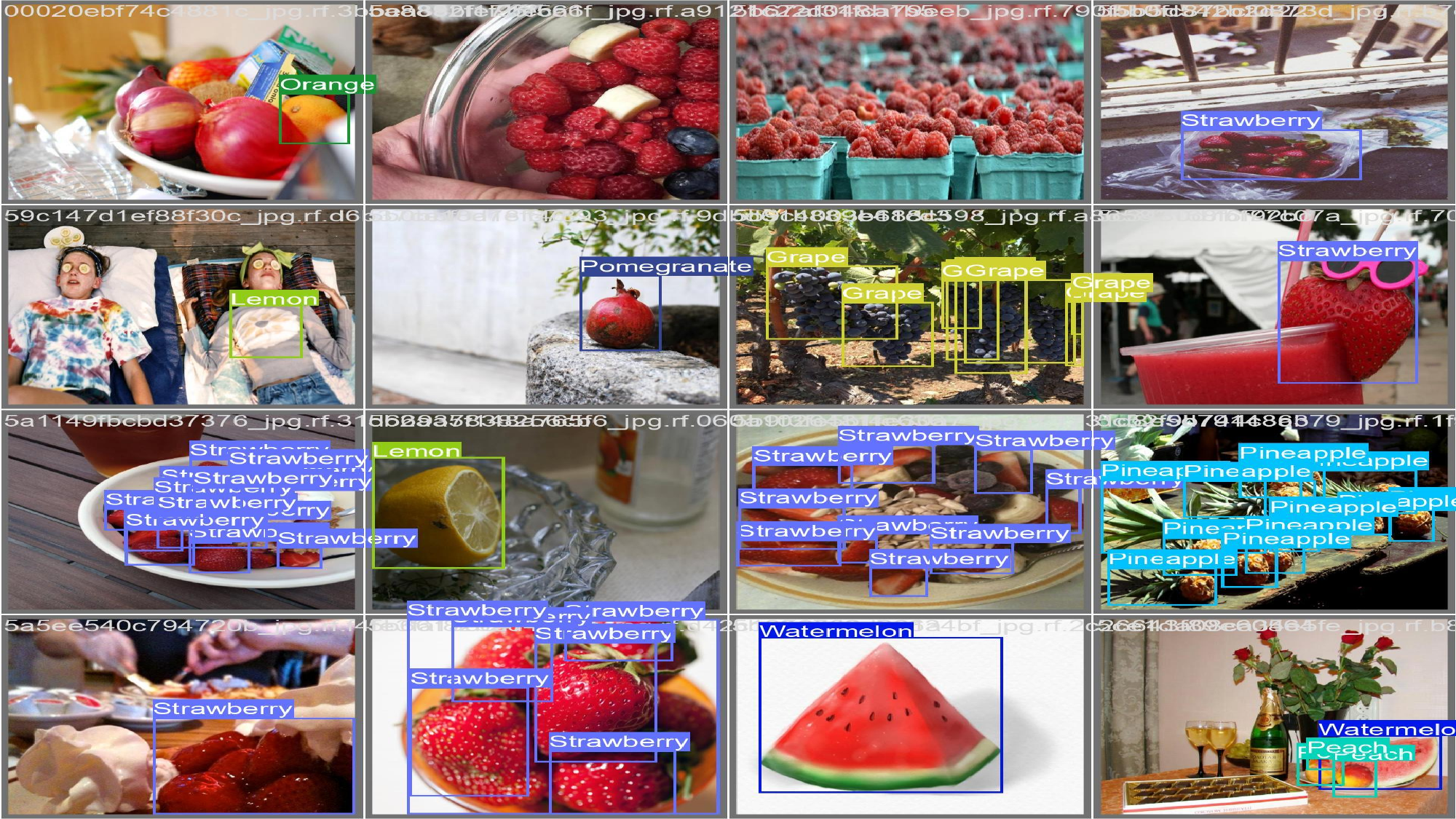}}
\caption{Fruit detection by FruitPAL 2.0}
\label{fig:allegic fruit2}
\end{figure}

Raspberry pi 4 with 8GB RAM and pi camera is used as Device Level as shows in Fig \ref{fig:FruitPAL 2.0}. In evaluating of "FruitPAL 2.0" model, the ability of FruitPAL 2.0 is showing in Fig.\ref{fig:allegic fruit2}. With high performance model and fast end device, FruitPAL 2.0 can encourage users to increase fruits consumption.

\begin{figure}[htbp]
\centerline{\includegraphics[width=14cm,height=25cm,keepaspectratio]{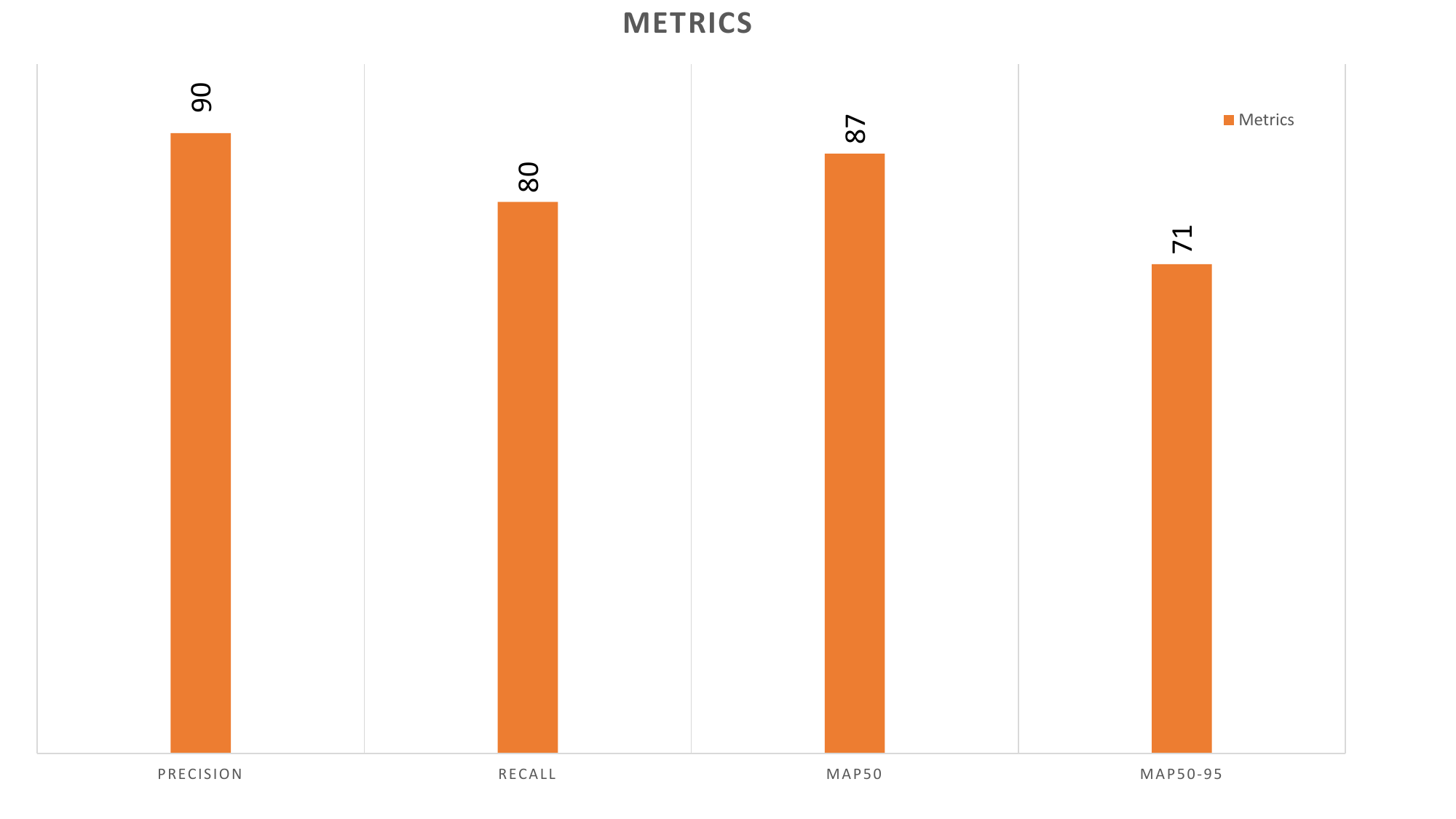}}
\caption{FruitPAL 2.0 Metrics}
\label{fig:Metrics}
\end{figure}

\begin{algorithm}
\caption{Process of Object Detection employ in FruitPAL 2.0 }
\begin{algorithmic}[1]
\State Time start  \Comment {the deivce in on }
\State Image is Captured.
\State Initial fruit list is created. 
    \For{Image is captured} {every hours}
        {
          \State Compare the new fruit list with initial fruit list.
         \If  {Fruit is eaten}
            {Save the eaten fruit in message list.
            
            }
         \ElsIf {New fruit add}
            {Update the initial fruit list.}
        \EndIf
      \State Add a hour in the timer.
        \If {Timer = 24 hours}
          {End For loop.}
         \EndIf

     }\EndFor
    
\State compare the message list with the table \ref{table:Value2}.
\State Message sends to the user

\end{algorithmic}
\end{algorithm}

\subsection{Dataset Details} \label{Dataset cp}

\subsubsection{Overview}
\begin{figure} [htbp]
\centerline{\includegraphics[width=14cm,height=25cm,keepaspectratio]{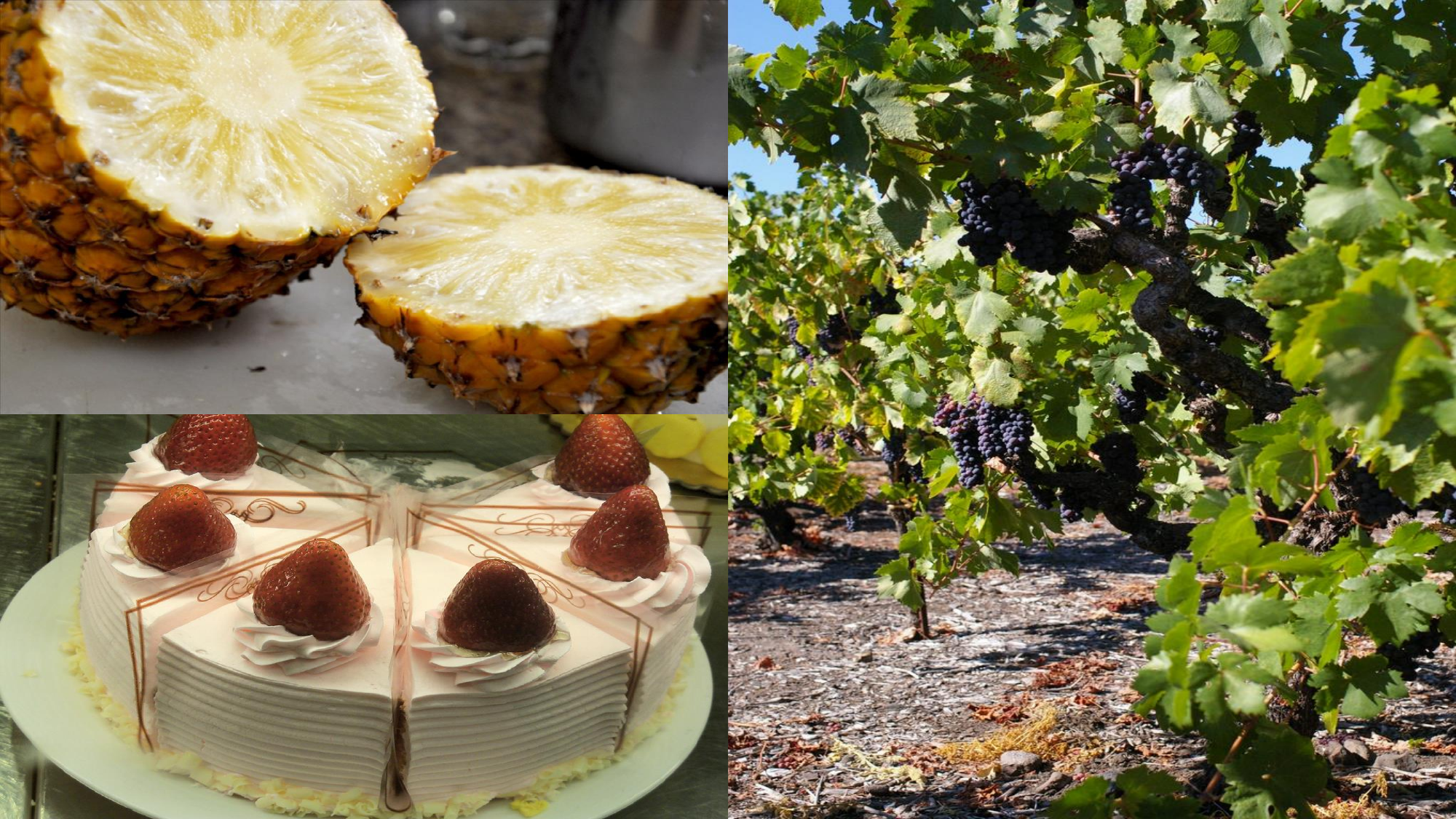}}
\caption{ Sample Images from the Allergic-fruit Dataset.}
\label{fig:Allergic-fruit}
\end{figure}

The quality of the dataset that is used in the object detection application can drive the quality of the results. FruitPAL and FruitPAL 2.0 devices use a specific dataset for fruit allergens that has been created by us \cite{Allergic-fruit}. Our dataset has around 3000 images that are collected from Open Images Dataset V7 \cite{datasetv7}. What makes our dataset distinguished in the fruit allergens field is that this dataset contains high quality images that have been captured in different environments, as shown in Fig \ref{fig:Allergic-fruit}. In addition, there are 15 classes  used in our dataset: Apple, Banana, Cantaloupe, Common fig, Grape, Grapefruit, Lemon, Mango, Orange, Peach, Pear, Pineapple, Pomegranate, Strawberry, and Watermelon.

\subsubsection{Annotation}

All classes in the Allergic-fruit dataset are annotated into different approaches based on:
\begin{itemize}
\item  Whole fruit.

\item  Cut fruit.

\item Fruit with rots.

\item  Fruit inside boxes.

\item  Fruit mixed with others. 
\end{itemize}

\begin{figure*}[htbp]
\centerline{\includegraphics[width=15cm,height=25cm,keepaspectratio]{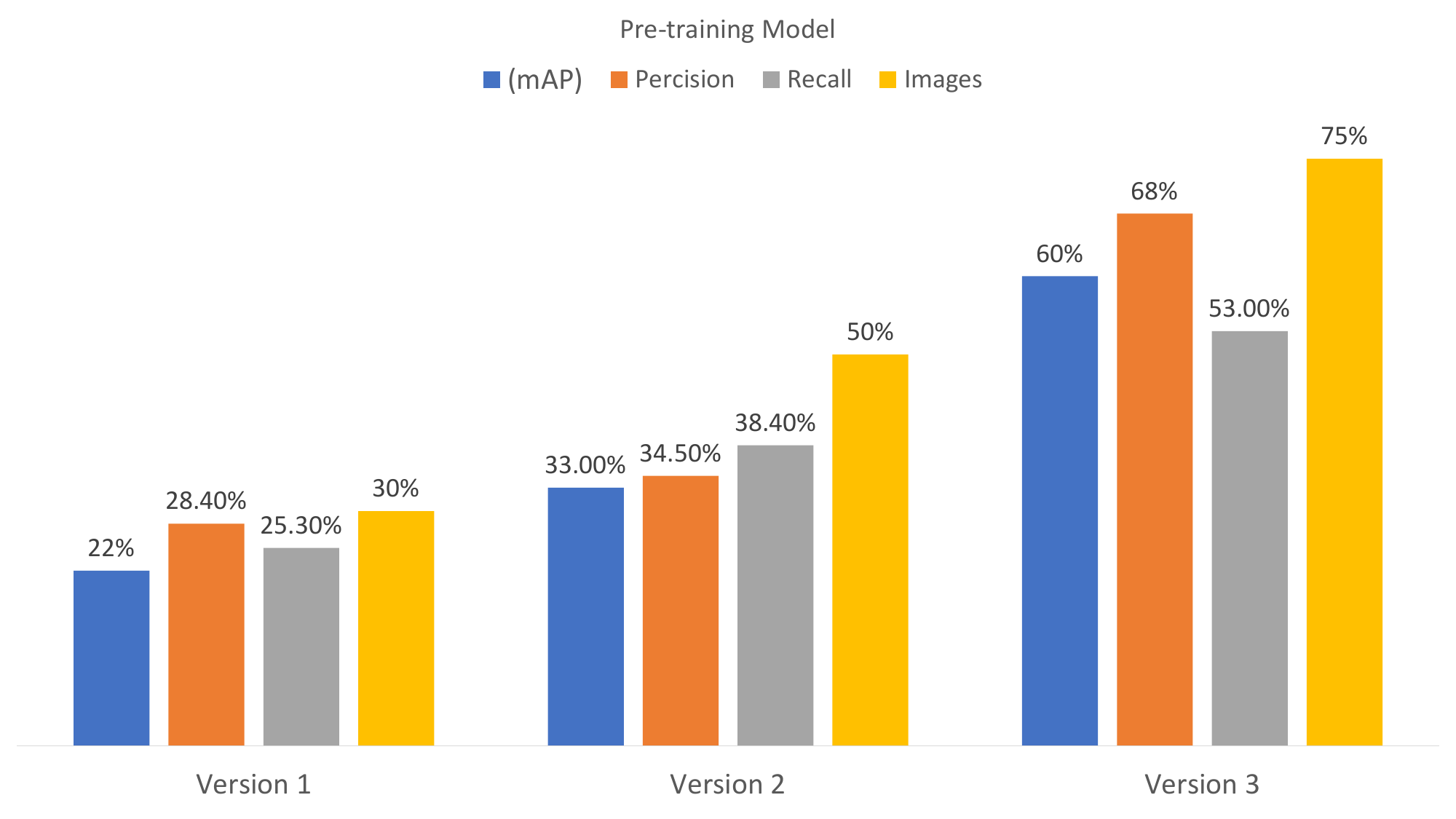}}
\caption{Pre-trained Model used to annotat the allergic dataset}
\label{fig:Pre-training Model}
\end{figure*}

When annotating images from this dataset, we started by using the predictions generated by a model that was trained with Roboflow Train. Pre-training the model is an proficient approach employed for annotating fruits in the allergic-fruit dataset. At the beginning of creating the dataset, the MS COCO model which has 55.8\% mAP is used. However, the MS COCO model has only three different kinds of fruit which are apple, banana, and orange. Three versions of the pre-trained model have been built at our end. The first version was trained based on the pictures annotated by the MS COCO model. The second and third versions relied on the previous versions' pre-trained model. Fig. \ref{fig:Pre-training Model} illustrates that each version's improvement mAP, precision, and recall were dominated by the number of annotated images in each version of the allergic-dataset. 

The Dataset Health Check is analyzed in Table \ref{table:Annotations}. A total of 16,000 boundary boxes have been drawn across 16 different classes, resulting in an average of 5.3 objects per image. Table \ref{table:balanc} shows the boundary boxes for each class in terms of training, validation, and testing.

\begin{scriptsize}
\begin{table}[htbp]
 \caption{Annotations}
  \centering
  \begin{tabular}{|l|l|l|}
    \hline
    \textbf{Class} & \textbf{Number of image}& \textbf{annotations}\\
    \hline
    \hline
    Strawberry & 630 & 3,368  \\
    \hline
    Orange & 519 & 2,957\\
    \hline
     Lemon & 426 & 1,638 \\
    \hline
    Pear & 149 & 547\\
    \hline 
    Pineapple & 212 & 539\\
    \hline 
    Grapefruit & 92 & 429\\
    \hline
    Peach & 138 & 1,425\\
    \hline
    Banana& 471 & 1,401\\
    \hline
    Common fig & 66 & 356\\
    \hline
    Apple & 188 & 631\\
    \hline
    Grape & 454 & 1,546\\
    \hline
    Mango & 105 & 639 \\
    \hline
    Watermelon & 223 & 660\\
    \hline
    Pomegranate & 131 & 343\\
    \hline
    Cantaloupe & 85 & 283\\
    \hline
    Null & 170 & 0\\
    \hline
    \hline
    Total & 3,862 without Null images &16,762\\
    \hline
  \end{tabular}
  \label{table:Annotations}
 
\end{table}
\end{scriptsize}

\begin{scriptsize}
      \begin{table}[htbp]
   \caption{ Boundary boxes for each kinds of fruits include training, validation, and testing}
                           \centering
                           \begin{tabular}{|l|l|l|l|}
                                         \hline
                                         \textbf{Class} & \textbf{Training}& \textbf{Validation} & \textbf{Testing}\\
                                         \hline
                                         \hline
                                         Strawberry & 2,458 & 592 & 314 \\
                                         \hline
                                         Orange & 2,019 & 607 & 321\\
                                         \hline
                                         Lemon & 958 & 449 & 241 \\
                                         \hline
                                         Pear & 374 & 122 & 56\\
                                         \hline 
                                         Pineapple & 346 & 135& 56\\
                                         \hline 
                                         Grapefruit & 342 & 65 &29\\
                                         \hline
                                         Peach & 888 & 411 &128\\
                                         \hline
                                         Banana& 831 & 416 &181\\
                                         \hline
                                         Common fig & 66 & 25 &33\\
                                         \hline
                                         Apple & 397 & 122 &107\\
                                         \hline
                                         Grape & 1118 & 295 &135\\
                                         \hline
                                         Mango & 481 & 85 & 72\\
                                         \hline
                                         Watermelon & 423 & 122&130\\
                                         \hline
                                         Pomegranate & 230 & 89&29\\
                                         \hline
                                         Cantaloupe & 166 & 71 &48\\
                                         \hline
                           \end{tabular}
                           \label{table:balanc}
                           
              \end{table}
\end{scriptsize}

\subsubsection{Image Augmentation}
 Increasing the number of images that have been duplicated in the dataset has been done by a technology called image augmentation. a technique to increase the number of images used. Image augmentation, generating synthetic images to enhance the model \cite{image-augmentation}, has been used in our dataset. To increase the accuracy of the dataset, Image augmentation has been applied on the annotated images, which are:
\begin{itemize}
\item  Grayscale: apply to 3\% of images.

\item  Saturation: Between -5\% and +5\%.

\item Brightness:Between -10\% and +10\%.

\item Exposure: Between -10\% and +10\%.

\item  Blur: Up to 0.5px.

\item  Noise: Up to 1\% of pixels.

\item Mosaic: Applied.
\end{itemize}

\begin{figure*}[htbp]
\centerline{\includegraphics[width=15cm,height=25cm,keepaspectratio]{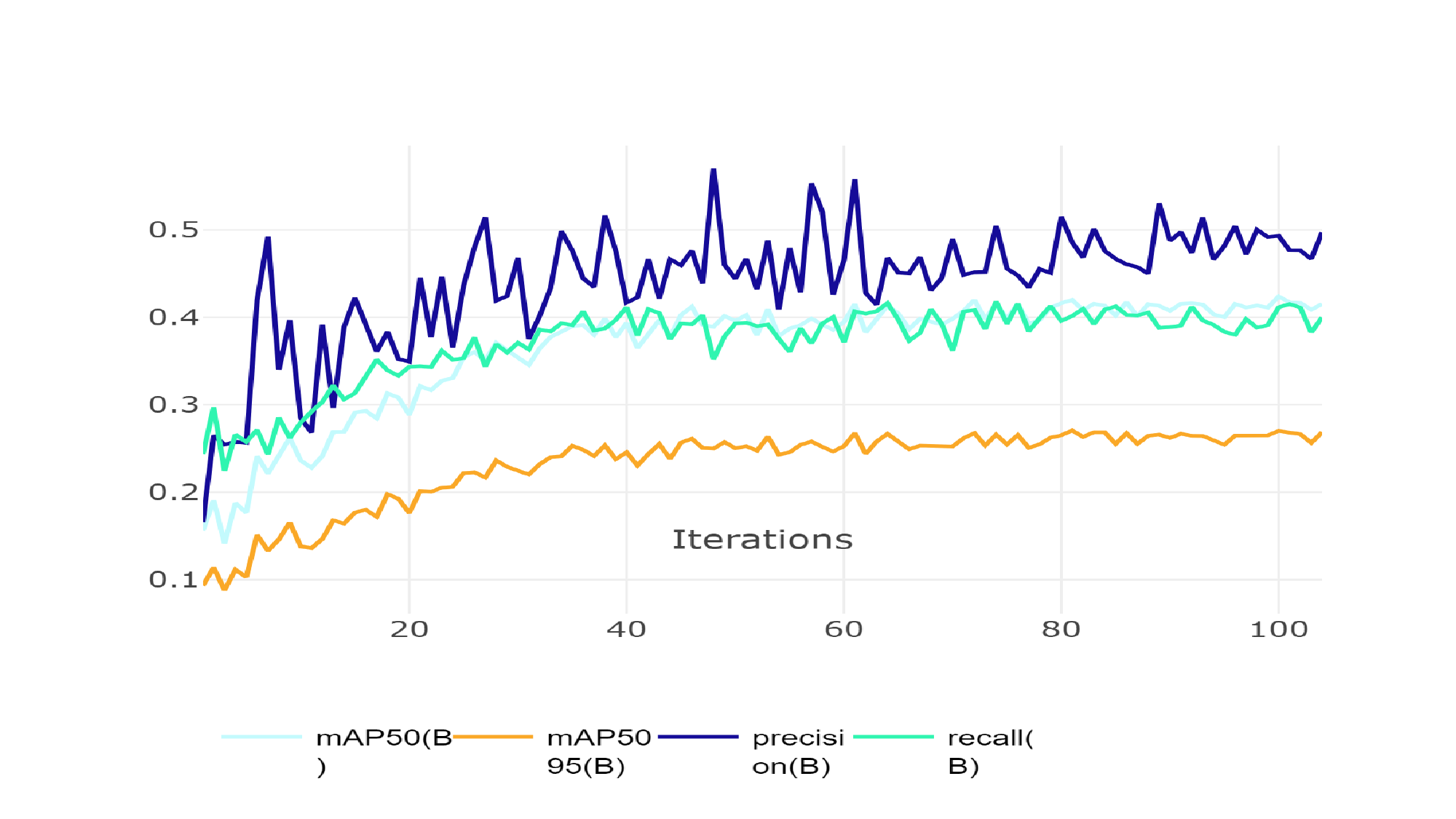}}
\caption{Evaluation of the Dataset Without Augmentation}
\label{fig:105 metrics}
\end{figure*}

The dataset has been evaluated with different sets of image augmentation on YOLOv8. The first evaluation without image augmentation, in Fig. \ref{fig:105 metrics} shows the metrics on 105 epochs, which are mAP50, mAP50-95, precise, and recall.
The second set of image augmentation is:
\begin{itemize}
\item  flip: horizontal, vertical

\item  Saturation: Between -25\% and +25\%.

\item  Noise: Up to 5\% of pixels.

\end{itemize}

\begin{figure*}[htbp]
\centerline{\includegraphics[width=13cm,height=25cm,keepaspectratio]{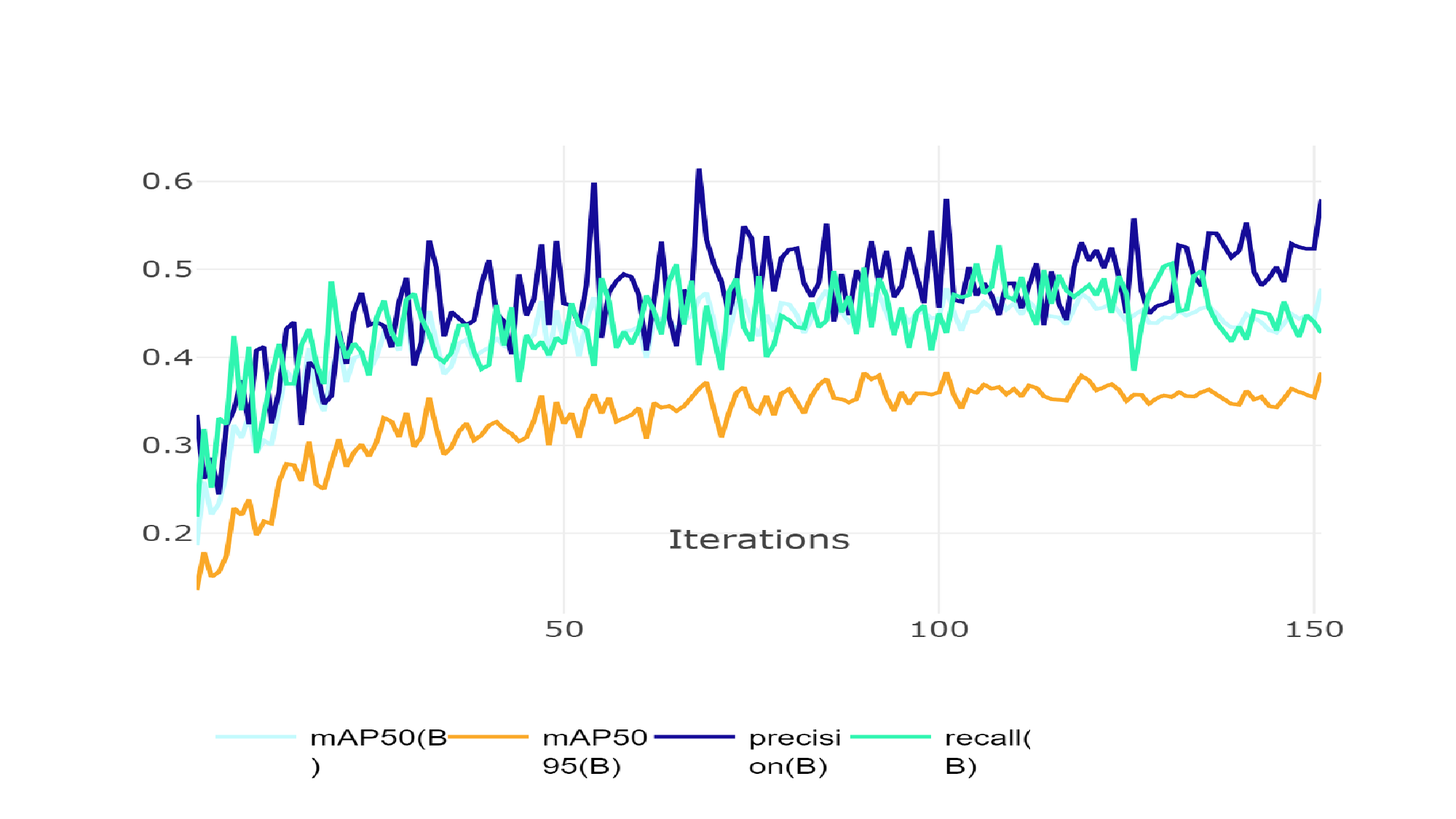}}
\caption{Evaluation of the Second set of Image Augmentation}
\label{fig:151 metrics}
\end{figure*}

\begin{figure*}[htbp]
\centerline{\includegraphics[width=13cm,height=25cm,keepaspectratio]{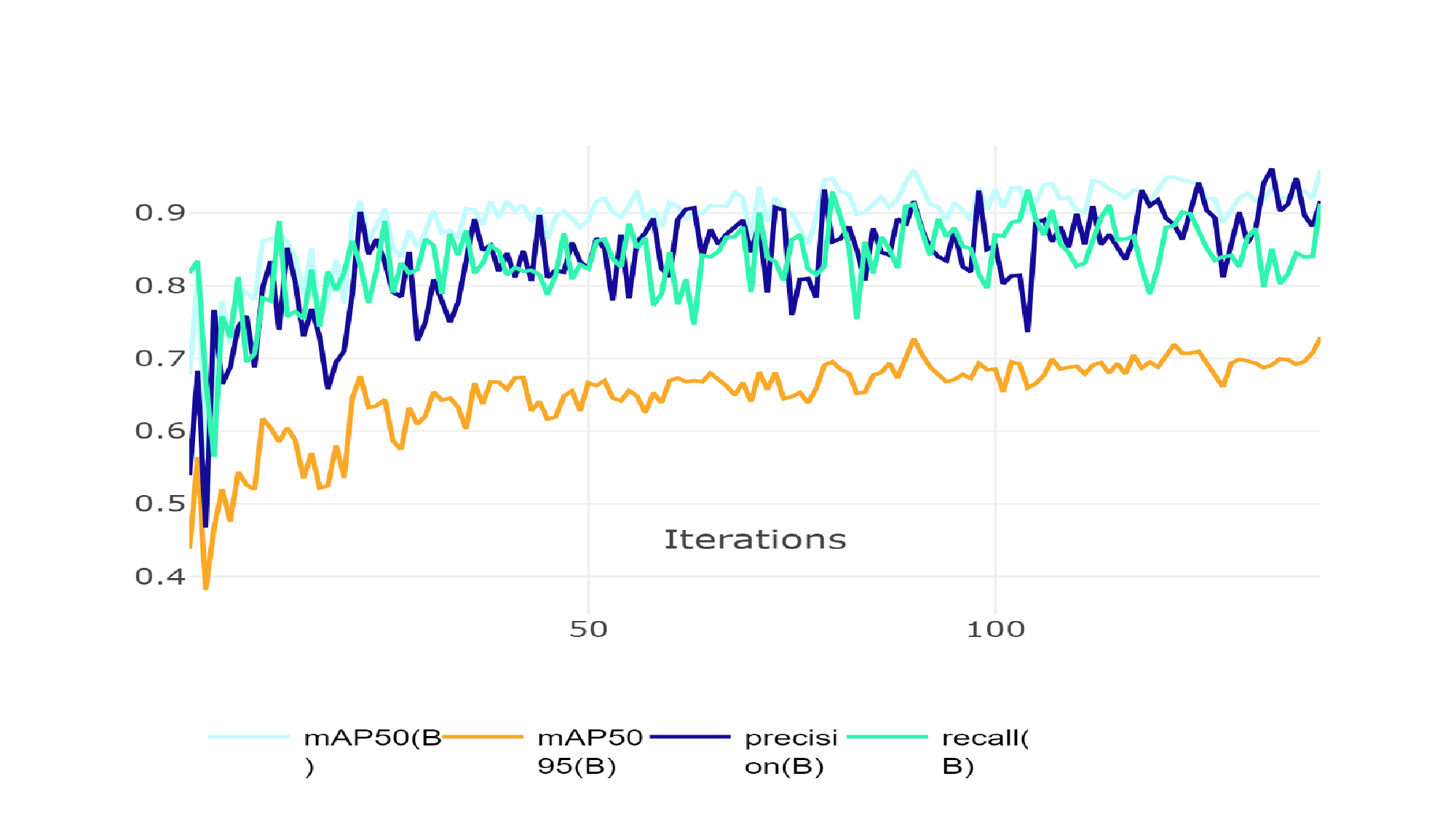}}
\caption{Evaluation of the third set of Image Augmentation}
\label{fig:140 metrics}
\end{figure*}

Figs. \ref{fig:151 metrics} shows the evaluation on 151 epochs. The model training was automatically stopped because there is no improvement over the epochs. Choosing the right augmentation plays a critical role in evaluation. The third set of image augmentation that was evaluated is shown in Figs. \ref{fig:140 metrics} is the following:
\begin{itemize}
\item  Grayscale: Apply to 3\% of images.

\item  Saturation: Between -5\% and +5\%.

\item Brightness: Between -10\% and +10\%.

\item Exposure: Between -10\% and +10\%. 

\item Blur: Up to 0.5 px.

\item  Noise: Up to 1\% of pixels.

\item  Mosaic: Applied.
\end{itemize}

\begin{figure*}[htbp]
\centerline{\includegraphics[width=15cm,height=25cm,keepaspectratio]{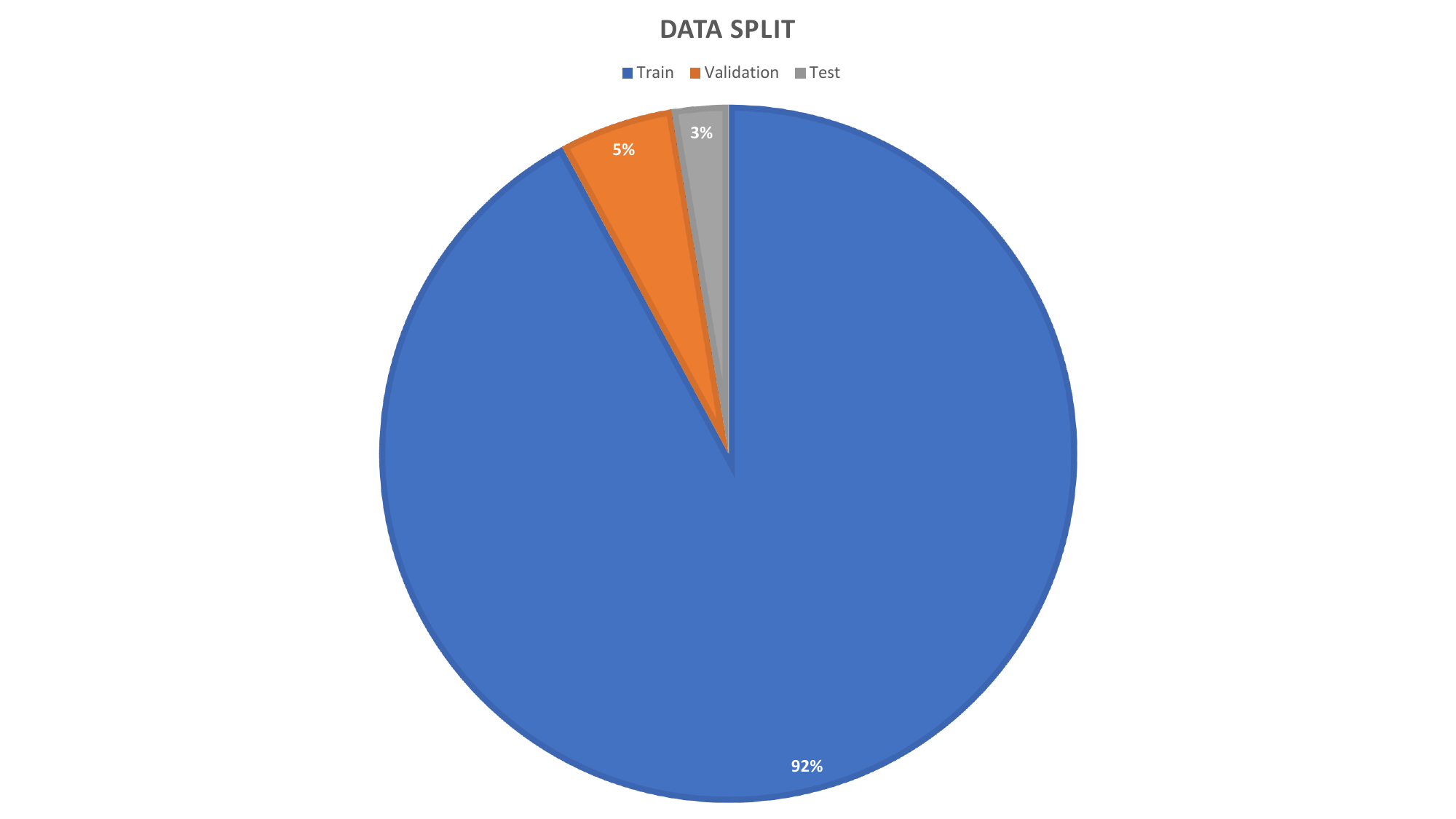}}
\caption{Data Split}
\label{fig:Data Split}
\end{figure*}

\begin{figure*}[htbp]
\centerline{\includegraphics[width=14cm,height=25cm,keepaspectratio]{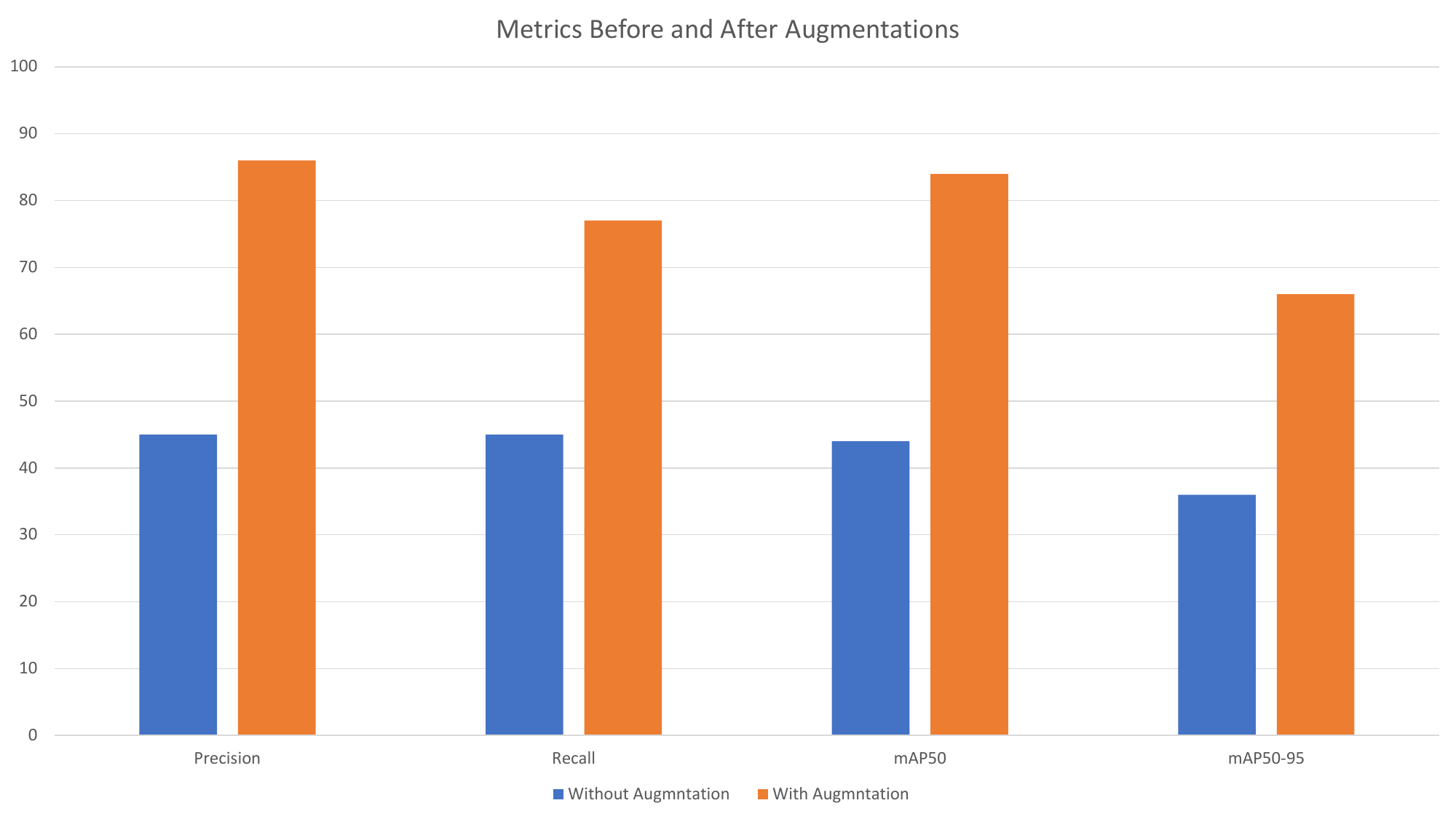}}
\caption{ Precision, recall, mAP50, and mAP50-95 With and Without Augmentations}
\label{fig:Augmentations}
\end{figure*}

Image augmentation increased the number of the images in the dataset to 12,061 images. The images were split between Training set, Validation set, and Testing set ,as shown in Fig. \ref{fig:Data Split}. The YOLOv8 model was utilized to evaluate the dataset in order to examine the impact of augmentation on the metrics, as shown in Fig. \ref{fig:Augmentations}.

\subsubsection{Evaluation}

The dataset was evaluated on YOLOv8m \ref{Yolov8m} and YOLOv5m V6.0 \ref{Yolov5m}. Table \ref{table:Evaluation dataset} displays evaluation of the Allergic-fruit dataset. The dataset exhibits a notable level of image quality, with annotations that demonstrate a high degree of accuracy. Table \ref{table:mAP50} presents the YOLOv8m and YOLOv5m V 6.0 mAP50 values for each class in the dataset.

\begin{scriptsize}
\begin{table}[htbp]
 \caption{Evaluation Dataset}
  \centering
  \begin{tabular}{|l|l|l|}
    \hline
    \textbf{Metrics} & \textbf{YOLOv8m} & \textbf{YOLOv5m V6.0}\\
    \hline
    \hline
    Precision & 86\% & 90\% \\
    \hline
    Recall & 77\% & 80\% \\
    \hline
     mAP50 & 84\% & 87\% \\
    \hline
    mAP50-95 & 66\% & 71\% \\
    \hline 
    Parameters & 25m & \\
    \hline
  \end{tabular}
  \label{table:Evaluation dataset}
 
\end{table}
\end{scriptsize}

\begin{scriptsize}
\begin{table}[htbp]
 \caption{mAP50 for Each Class on Allergic-fruit dataset}
  \centering
  \begin{tabular}{|l|l|l|}
    \hline
    \textbf{Class} & \textbf{YOLOv8}& \textbf{YOLOv5m V6.0}\\
    \hline
    \hline
    Strawberry & 87.5\% & 89.6\%  \\
    \hline
    Orange & 81.6\% & 82.2\%\\
    \hline
     Lemon & 78.8\% & 83.0\%\\
    \hline
    Pear & 88.0\% & 91.6\%\\
    \hline 
    Pineapple & 85.5\% & 88.7\%\\
    \hline 
    Grapefruit & 82.1\% & 81.5\%\\
    \hline
    Peach & 94.1\% & 95.8\%\\
    \hline
    Banana& 76.3\% & 77.0\%\\
    \hline
    Common fig & 89.7\% & 90.6\%\\
    \hline
    Apple & 59.1\% & 64.2\%\\
    \hline
    Grape & 76.2\% & 82.7\%\\
    \hline
    Mango & 94.8\% & 96.3\%\\
    \hline
    Watermelon & 93.6\% & 94.4\%\\
    \hline
    Pomegranate & 94.1\% & 98.0\%\\
    \hline
    Cantaloupe & 92.5\% & 91.4\%\\
    \hline
    \hline
    mAP50 & 84.9\% &87.1\%\\
    \hline
  \end{tabular}
  \label{table:mAP50}
 
\end{table}
\end{scriptsize}

\section{CONCLUSION}
The dataset is the primary asset in boosting the effectiveness of the Object detection model. Most the fruit dataset has few kinds of fruit , so we build Allergic-fruit dataset that consists of fifteen different kinds of fruit. We employed different ways to annotate the dataset. The high efficiency methodology is employed for the annotation of the collection image. We conducted several levels of dataset evaluation. We made changes on different factors. On the annotated images, image augmentation has been applied to improve the accuracy of the dataset. We utilized the fruit object detection dataset to identify fruit allergies in FruitPAL A novel device for fruit allergens detection has been presented. Fruit allergens could cause significant harm for allergic people. Research discusses different ways to protect the humans from the allergens \cite{FrameworktoIdentifyAllergen, seasonalallergic, asthma}. We concentrated the group of people who lack recognition about fruit allergen. We chose the faster time respond and high accuracy model which is YOLOv8. Another use of our dataset is FruitPAL2.0, a methodology that offers an automated system for monitoring fruit consumption. We focused on an important problem: people who don't eat enough fruit. Consuming fruits contributes to the enhancement of overall health through the provision of essential nutrients \cite{SugarContent,nutrition,Soluble,HealthyFruits}. Nutritionists encourage the consumption of fruits due to their potential role in preventing diseases\cite{Incredible}.The proposed work is fully automatic device that can monitor the fruit consumption. Two versions of the YOLO family were used in this work to solve two main problems that had to do with fruit. We also used different models to test our dataset.

As for future research purposes, increasing the number of classes on the dataset can be achieved by collecting images. In the event that a caregiver fails to intervene in order to protect an allergic person, it is essential that the system promptly initiate contact with emergency services and transmit the precise location of the affected individual in terms of heath protection. Additionally, the systematic can provide more security since our world exposes to illness and other differ types of allergies. Furthermore, the inclusion of a speaker and microphone allows communication between the caregiver and the allergy sufferer. Also, caregiver can lead the allergic person through out the communication channel to avoid consuming the allergen fruit, so the allergic person can eat other heather fruits. Creating an application that allows the caregiver to observe the allergic person in real-time is an additional feature to be considered for the future. Future research should concentrate on including face recognition since it can increase efficiency. The implementation of face recognition technology on FruitPAL enables the device to accurately identify individuals and fruit they should not eat. FruitPAL 2.0 can be used by multiple users due to face recognition. Connecting the health record with the FruitPAL 2.0 cloud platform has a chance to enhance the system's ability to encourage users to consume specific fruits that are beneficial to their health. Family doctors can identify the fruits which may affect the allergic patients. One type of food that cause choking is fruits because some fruits are large, solid fruit which may be hard to swallow easily. The device can identify the choking situations and alert the physician to protect human lives.

\bibliography{Bibliography_FruitPAL}

\begin{wrapfigure}{l}{0.25\textwidth}
    \centering
    \includegraphics[width=0.85 in,height=1.25in,clip,keepaspectratio]{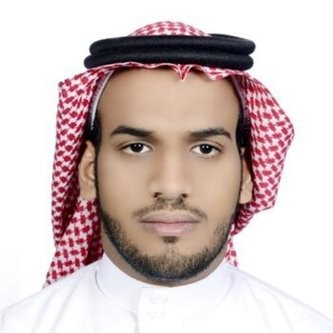} 
\end{wrapfigure}
\textbf{Abdulrahman I. Alkinani} received a bachelor’s degree (Honors) in computer engineering from Al Baha University, Saudi Arabia-Albaha, in 2017 and a master’s in computer engineering in 2023 from University of North Texa, Texas-Denton. USA. A Faculty Member in University of Tabuk, Saudi Arabia-Tabuk at the department of Computer Engineering. Currently a Ph.D. Student in the research group at Smart Electronics Systems Laboratory (SESL) at Computer Science and Engineering at the University of North Texas, Denton, TX. Research interests include IoT, Cyber Physical Systems (CPS), Artificial Intelligence (AI) in Smart Healthcare.

\begin{wrapfigure}{l}{0.25\textwidth}
    \centering
    \includegraphics[width=1in,height=1.25in,clip,keepaspectratio]{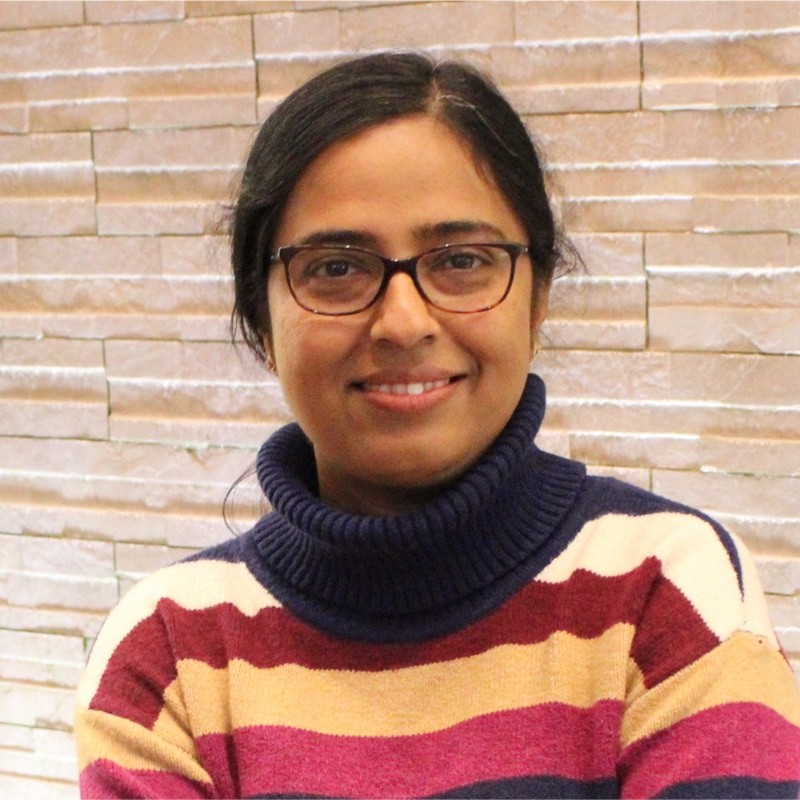} 
\end{wrapfigure}
\textbf{Alakananda Mitra} is a Research Assistant Professor at the Nebraska Water Center, University of Nebraska-Lincoln, NE, USA and a visiting computer scientist at the Adaptive Cropping Systems Laboratory, USDA-ARS, Beltsville, MD, USA. She earned her Ph.D. in computer science and engineering from the University of North Texas, Denton, TX, USA. Dr. Mitra earned her Bachelor of Science degree (Hons.) in physics from the Presidency College, University of Calcutta, and her B. Tech. and M. Tech. degrees in radiophysics and electronics from the Institute of Radiophysics and Electronics, University of Calcutta, India. 
Her research focuses on "Precision in Practice: Application Specific AI". She aims to solve real-world problems using AI and ML techniques. Her work spans various fields, with a focus on cybersecurity, smart agriculture, and smart healthcare. Dr. Mitra strives to make AI solutions widely available, affordable, and accessible to people. Her areas of expertise include AI, machine learning, deep learning, computer vision, and edge computing. Currently, she is working on AI-based crop models, tinyML devices for plant disease detection, and the application of federated learning in smart agriculture. Dr. Mitra is also developing a crop and soil simulation model, databases, and other suitable agro-climatology modeling tools. Notably, during the course of her doctoral research, she was honored with the Outstanding Doctoral Student Award, in addition to receiving several Best Paper awards. Furthermore, Dr. Mitra holds two pending US patents.

\begin{wrapfigure}{l}{0.25\textwidth}
    \centering
    \includegraphics[width=1in,height=1.25in,clip,keepaspectratio]{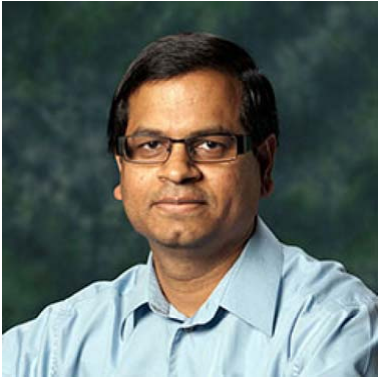} 
\end{wrapfigure}
\textbf{Saraju P. Mohanty} received the bachelor’s degree (Honors) in electrical engineering from the Orissa University of Agriculture and Technology, Bhubaneswar, in 1995, the master’s degree in Systems Science and Automation from the Indian Institute of Science, Bengaluru, in 1999, and the Ph.D. degree in Computer Science and Engineering from the University of South Florida, Tampa, in 2003. He is a Professor with the University of North Texas. His research is in “Smart Electronic Systems” which has been funded by National Science Foundations (NSF), Semiconductor Research Corporation (SRC), U.S. Air Force, IUSSTF, and Mission Innovation. He has authored 450 research articles, 5 books, and 9 granted and pending patents. His Google Scholar h-index is 54 and i10-index is 227 with 12,000 citations. He is regarded as a visionary researcher on Smart Cities technology in which his research deals with security and energy aware, and AI/ML-integrated smart components. He introduced the Secure Digital Camera (SDC) in 2004 with built-in security features designed using Hardware Assisted Security (HAS) or Security by Design (SbD) principle. He is widely credited as the designer for the first digital watermarking chip in 2004 and first the low-power digital watermarking chip in 2006. He is a recipient of 16 best paper awards, Fulbright Specialist Award in 2020, IEEE Consumer Electronics Society Outstanding Service Award in 2020, the IEEE-CS-TCVLSI Distinguished Leadership Award in 2018, and the PROSE Award for Best Textbook in Physical Sciences and Mathematics category in 2016. He has delivered 22 keynotes and served on 14 panels at various International Conferences. He has been serving on the editorial board of several peer-reviewed international transactions/journals, including IEEE Transactions on Big Data (TBD), IEEE Transactions on Computer-Aided Design of Integrated Circuits and Systems (TCAD), IEEE Transactions on Consumer Electronics (TCE), and ACM Journal on Emerging Technologies in Computing Systems (JETC). He has been the Editor-in-Chief (EiC) of the IEEE Consumer Electronics Magazine (MCE) during 2016-2021. He served as the Chair of Technical Committee on Very Large Scale Integration (TCVLSI), IEEE Computer Society (IEEE-CS) during 2014-2018 and on the Board of Governors of the IEEE Consumer Electronics Society during 2019-2021. He serves on the steering, organizing, and program committees of several international conferences. He is the steering committee chair/vice-chair for the IEEE International Symposium on Smart Electronic Systems (IEEE-iSES), the IEEE-CS Symposium on VLSI (ISVLSI), and the OITS International Conference on Information Technology (OCIT). He has mentored 2 post-doctoral researchers, and supervised 15 Ph.D. dissertations, 26 M.S. theses, and 20 undergraduate projects.

\begin{wrapfigure}{l}{0.25\textwidth}
    \centering
    \includegraphics[width=1in,height=1.25in,clip,keepaspectratio]{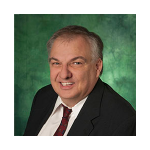} 
\end{wrapfigure}
\textbf{Elias Kougianos} received a BSEE from the University of Patras, Greece in 1985 and an MSEE in 1987, an MS in Physics in 1988 and a Ph.D. in EE in 1997, all from Louisiana State University. From 1988 through 1998 he was with Texas Instruments, Inc., in Houston and Dallas, TX. In 1998 he joined Avant! Corp. (now Synopsys) in Phoenix, AZ as a Senior Applications engineer and in 2000 he joined Cadence Design Systems, Inc., in Dallas, TX as a Senior Architect in Analog/Mixed-Signal Custom IC design. He has been at UNT since 2004. He is a Professor in the Department of Electrical Engineering, at the University of North Texas (UNT), Denton, TX. His research interests are in the area of Analog/Mixed-Signal/RF IC design and simulation and in the development of VLSI architectures for multimedia applications. He is an author of over 200 peer-reviewed journal and conference publications.

\end{document}